\documentclass[11pt]{article}

\usepackage[final]{acl}

\usepackage{times}
\usepackage{latexsym}

\usepackage[T1]{fontenc}

\usepackage[utf8]{inputenc}

\usepackage{microtype}

\usepackage{inconsolata}

\usepackage{graphicx}

\usepackage{times,latexsym}
\usepackage{url}
\usepackage[T1]{fontenc}
\usepackage{microtype}
\usepackage{booktabs}
\usepackage{amsmath}
\usepackage{amssymb}
\usepackage{pifont}
\usepackage{bbm}
\usepackage{array}
\usepackage{xspace}

\newcommand{\dataset}{\textsc{HealthDial}\xspace}
\definecolor{ourred}{HTML}{E13342}
\definecolor{ourblue}{HTML}{6495ed}

\newcommand{\rparagraph}[1]{\vspace{1.6mm}\noindent\textbf{#1.}}

\newcommand{\rrparagraph}[1]{\vspace{0.5mm}\noindent\textit{#1:}}

\newcommand{\eng}{{\textsc{eng}}\xspace}
\newcommand{\ara}{{\textsc{ara}}\xspace}
\newcommand{\spa}{{\textsc{spa}}\xspace}
\newcommand{\zho}{{\textsc{zho}}\xspace}

\newcommand{\lan}{{\textsc{lan}}\xspace}

\newcommand{\bluecheck}{{\color{racing-green}{\pmb{\checkmark}}}}
\newcommand{\xmark}{\color{awesome-red}\ding{55}}%

\definecolor{Gray}{gray}{0.92}
\definecolor{racing-green}{rgb}{0.0, 0.8, 0.6}
\definecolor{awesome-red}{rgb}{1.0, 0.13, 0.32}
\newcolumntype{Y}{>{\centering\arraybackslash}X}

\usepackage{todonotes}
\makeatletter
\newcommand*\iftodonotes{\if@todonotes@disabled\expandafter\@secondoftwo\else\expandafter\@firstoftwo\fi}
\makeatother

\usepackage[inline]{enumitem}
\usepackage[table,xcdraw]{xcolor}

\usepackage{tabularx}
\usepackage{subcaption}

\usepackage{multirow} 
\usepackage{arydshln}

\usepackage{listings}
\usepackage{xcolor}

\definecolor{stringcolor}{rgb}{0.1,0.5,0.1}
\definecolor{keycolor}{rgb}{0.0,0.0,0.6}
\definecolor{valuecolor}{rgb}{0.6,0.1,0.1}
\definecolor{bggray}{gray}{0.97}
\definecolor{commentgray}{gray}{0.5}

\lstdefinelanguage{json}{
    basicstyle=\ttfamily\small,
    numbers=none,
    backgroundcolor=\color{bggray},
    stepnumber=1,
    showstringspaces=false,
    breaklines=true,
    frame=single,
    rulecolor=\color{gray},
    string=[s]{"}{"},
    stringstyle=\color{stringcolor},
    identifierstyle=\color{keycolor},
    keywordstyle=\color{valuecolor},
    morestring=[b]',
    literate=
     *{:}{{{\color{black}:}}}{1}
      {,}{{{\color{black},}}}{1}
      {[}{{{\color{black}[}}}{1}
      {]}{{{\color{black}]}}}{1}
      {\{}{{{\color{black}\{}}}{1}
      {\}}{{{\color{black}\}}}}{1},
}

\usepackage[most]{tcolorbox}
\tcbuselibrary{listings, skins}

\newtcolorbox[use counter=lstlisting]{examplebox}[2][]{%
  colback=gray!20,
  colframe=black,
  width=\linewidth,
  boxsep=5pt,
  left=2pt,
  right=2pt,
  top=2pt,
  bottom=2pt,
  fonttitle=\bfseries,
  title={Example~\thelstlisting: #2},
  label={#1},
  enhanced,
  breakable,
  }

\title{\textit{Dial \dataset for Advice}: A Multilingual and Multi-Parallel Spoken Dialogue Dataset for Knowledge-Grounded Information Seeking}

\author{
  Songbo Hu$^{1}$\thanks{~~Equal contribution.}~~~ 
  Yinhong Liu$^{1}$\footnotemark[1]~~~
  Ej Zhou$^{1}$\footnotemark[1]~~\\
  \textbf{Evgeniia Razumovskaia$^{1}$~~~
  Xiaobin Wang$^{2}$~~~
  Alexander Fraser$^{3}$}\\
  \textbf{Ivan Vuli\'{c}$^{1}$\thanks{~~Equal senior contribution.}~~~
  Anna Korhonen$^{1}$\footnotemark[2]}
  \\
  $^{1}$Language Technology Lab, University of Cambridge, UK
  \\
  $^{2}$Independent Researcher
  \\
  $^{3}$School of Computation, Information and Technology, Technical University of Munich, Germany
  \\
  $^{1}$\texttt{\{sh2091, yl535, yz926, iv250, alk23\}@cam.ac.uk} \\
  $^{1}$\texttt{evgeniiarazum@google.com}
  $^{2}$\texttt{wxb9585@gmail.com} 
}

\begin{document}
\maketitle

\begin{abstract}
Creating spoken dialogue datasets is methodologically challenging, and these challenges are amplified when the goal is to build multilingual, multi-parallel datasets at scale. This work introduces \dataset, a large-scale, multilingual, and multi-parallel dataset for developing and evaluating retrieval-augmented generation (RAG)–based spoken dialogue systems. The dataset comprises 6,000 information-seeking dialogues (1,500 per language) grounded in trusted content from the World Health Organization (WHO) and 163 hours of user speech recorded from native speakers of diverse dialects across four official WHO languages: Arabic, Chinese, English, and Spanish. Each speaker is annotated with demographic (e.g., gender, age) and sociolinguistic (e.g., primary language, region of origin) variables. We report benchmark results across key dialogue tasks, which reveal consistent performance disparities across languages, even among high-resource ones. To support future research, we release the dataset, a prototype system, and a toolkit for data collection and system evaluation. 
\end{abstract}

\section{Introduction}
\label{sec:introduction}

Despite being the primary medium of human communication, speech remains under-represented in dialogue system research. When speech is incorporated, it is typically processed through a modular pipeline: automatic speech recognition (ASR) converts speech to text, a text-based dialogue model generates a response, and text-to-speech (TTS) synthesises it back into audio~\cite{6407655}. While effective, this design normalises away important aspects of spoken language, such as accent, dialect, and sociolinguistic variation.

Speech-first dialogue datasets are essential for enabling research into fully speech-based dialogue pipelines and for benchmarking emerging speech-native language models. However, constructing spoken dialogue datasets is both methodologically complex~\cite[\textit{inter alia}]{hemphill-etal-1990-atis, kim2021robust, 10.5555/3666122.3667821} and ethically challenging~\cite{9102875,shahin2023}, due to the personally identifiable nature of speech signals. These difficulties are amplified in multilingual settings, especially when collecting multi-parallel datasets across languages~\cite{caswell-etal-2020-language}, as spontaneous parallel dialogues rarely occur naturally~\cite{bawden2021diabla, goncalo-oliveira-etal-2022-brief}.

In this work, we present a large-scale data collection process for constructing multilingual, multi-parallel spoken dialogues. We take a bottom-up, outline-based approach~\cite{Majewska:2023cod}, in which native speakers realise language-agnostic dialogue schemata, constructed with LLMs as high-level prompts for annotators, into naturalistic utterances in their respective languages. This design balances content control with linguistic diversity, while reducing privacy risks by relying on hypothetical rather than real user interactions.

The outcome of this data collection process is \dataset, which contains 6,000 knowledge-grounded, information-seeking dialogues across four WHO languages: Arabic, Chinese, English, and Spanish. In total, the dataset provides 163 hours of spoken user utterances recorded by native speakers from diverse language varieties, with each dialogue annotated with speaker demographics. Dialogue responses are grounded in a curated knowledge base of WHO health snippets.

In this paper, we elaborate on the key properties of \dataset and position it within the context of existing resources. We then present a large-scale data collection process behind the creation of the dataset. We establish benchmark results across multiple NLP tasks for all four languages: ASR, TTS, knowledge retrieval, and knowledge filtering. In addition, we present example analyses enabled by the dataset. These results clearly indicate the challenging nature of the dataset and reveal performance disparities across different languages under current models. 

\rparagraph{Code and Data}
We release the full dataset, baseline code for benchmarking, the prototype dialogue system, and a toolkit for replicating the data collection process and system evaluation: \href{https://github.com/cambridgeltl/healthdial}{\nolinkurl{github.com/cambridgeltl/healthdial}}.

\section{Related Work}
\label{sec:related_work}

\begin{table*}[t!]
\centering
\def\arraystretch{0.75}
{\scriptsize 
\resizebox{\textwidth}{!}{%
\begin{tabular}{l ccc ccccc}
\toprule
\rowcolor{Gray}
\textbf{Dataset (Reference)} & \# Langs & \# Dials & Domain & Info Seeking? & Knowledge? & Multi-P? & Speech? & Speaker Metadata? \\
\cmidrule(lr){2-4} \cmidrule(lr){5-9}
Fisher\&CALLHOME~\cite{post-etal-2013-improved} & 2 & 939 & general & {\bluecheck} & {\xmark} & {\bluecheck} & {\hspace{5pt}}{\bluecheck} & {\bluecheck} \\
TourSG (DSTC 5)~\cite{7846311} & 2 & 36 & tourism & {\bluecheck} & {\bluecheck} & {\xmark} & {\bluecheck} & {\xmark} \\
WOZ 2.0~\cite{mrksic-etal-2017-semantic} & 3 & 1000 & tourism & {\bluecheck} & {\bluecheck} & {\bluecheck} & {\xmark} & {\xmark} \\
MedDialog~\cite{zeng-etal-2020-meddialog} & 2 & 1.8m & health & {\bluecheck} & {\xmark} & {\xmark} & {\xmark} & {\bluecheck} \\
BiToD~\cite{Lin:2021bitod} & 2 & 3345 & tourism & {\bluecheck} & {\bluecheck} & {\xmark} & {\xmark} & {\xmark} \\
AllWOZ~\cite{Zuo:2021allwoz} & 8 & 90 & tourism & {\bluecheck} & {\bluecheck} & {\bluecheck} & {\xmark} & {\xmark} \\
XPersona~\cite{lin-etal-2021-xpersona} & 7 & 556 & general & {\xmark} & {\xmark} & {\bluecheck} & {\xmark} & {\xmark} \\
GlobalWOZ~\cite{ding-etal-2022-globalwoz} & 21 & 500 & tourism & {\bluecheck} & {\bluecheck} & {\xmark} & {\xmark} & {\xmark} \\
Multi2WOZ~\cite{hung-etal-2022-multi2woz} & 5 & 1000 & tourism & {\bluecheck} & {\bluecheck} & {\bluecheck} & {\xmark} & {\xmark} \\
Multi3WOZ~\cite{hu-etal-2023-multi-3} & 4 & 8300 & tourism & {\bluecheck} & {\bluecheck} & {\bluecheck} & {\xmark} & {\xmark} \\
XDailyDialog~\cite{liu-etal-2023-xdailydialog} & 4 & 1300 & tourism & {\xmark} & {\xmark} & {\bluecheck} & {\xmark} & {\xmark} \\
SpeechBSD~\cite{shimizu-etal-2023-towards} & 2 & 808 & business & {\xmark} & {\xmark} & {\bluecheck} & {\xmark} & {\bluecheck} \\
\cmidrule(lr){2-4} \cmidrule(lr){5-9}
\textbf{\dataset (this work)} & 4 & 1500 & health & {\bluecheck} & {\bluecheck} & {\bluecheck} & {\bluecheck} & {\bluecheck} \\
\bottomrule
\end{tabular}
}
}

\caption{Summary of multilingual dialogue datasets. Datasets are included based on the following criteria: (i) support for multiple languages, (ii) provision of multi-turn interactions, and (iii) public availability with scientific publications. \textbf{\# Langs} refers to the number of supported languages (including English). \textbf{\# Dials} refers to the average number of human-authored or human-curated dialogues per language. \textbf{Info Seeking} denotes whether dialogues are task-oriented and involve information-seeking scenarios, as opposed to casual chitchat. \textbf{Knowledge} indicates whether dialogue turns are grounded in external knowledge sources. \textbf{Multi-P} refers to the multi-parallelism of dialogues in the dataset. \textbf{Speech} refers to the availability of spoken modality. \textbf{Speaker Metadata} indicates the annotation of the speaker's demographic and sociolinguistic background.}

\label{tab:dialogue_summary_1}
\end{table*}

We now delve deeper into the main benefits of \dataset, characterising how its key properties make it a unique language resource. The summary and statistics of the most relevant prior work on multilingual dialogue datasets are provided in Table~\ref{tab:dialogue_summary_1}. Building upon this table, we discuss those dialogue datasets along with other related work in what follows, focusing on the four desirable properties of \dataset and how these counteract the detected main limitations of other datasets.

\vspace{0.5mm}
\noindent \textbf{P1. Information-Seeking Dialogues in Multiple Languages and Speech.}
There has been a growing interest in creating datasets to mitigate the language resource gap in multilingual NLP~\cite{ponti-etal-2019-modeling,joshi-etal-2020-state}. However, this gap remains particularly pronounced in the domain of dialogue, with only a handful of datasets offering multi-turn dialogues in multiple languages. The scarcity is even more pronounced for spoken dialogues with only one dataset providing on average 36 spoken dialogues per language across Chinese and English~\cite{7846311}. To the best of our knowledge, \dataset is the first large-scale multilingual dialogue dataset that includes both speech and text across four languages.

Beyond multilingual dialogue datasets, other closely related resources include speech translation datasets~\cite{federmann-lewis-2016-microsoft,federmann-lewis-2017-microsoft,jia-etal-2022-cvss, le2025multimed} and natural language understanding (NLU) datasets (see the survey by~\citet{Razumovskaia:2022survey}). While these resources may include utterances in spoken form, they typically present them `in isolation', lacking the essential features of multi-turn interaction. Another line of related work is code-switching dialogue datasets~\cite{deuchar2010bilingbank,ramanarayanan2017jee}, which support multi-turn dialogues but contain only isolated phrases from a secondary language. \dataset addresses these limitations by offering a parallel multilingual dialogue dataset, with an equal number of dialogues across four languages.

\vspace{0.5mm}
\noindent \textbf{P2. Knowledge-Grounded Dialogues in the Health Domain.}
Existing dialogue datasets in the health domain are typically sourced from online medical forums or consultation transcripts~\cite{zeng-etal-2020-meddialog, 10.1145/3404835.3462921, he2022dialmed, liu2022meddg}. While these datasets capture realistic patient-doctor interactions, they 
have several limitations:
(i) most contain only text-based consultations;
(ii) they are available exclusively in Chinese or English, whereas the populations most in need are often speakers of other lower-resourced languages~\cite{worldbank2023digital, rodriguez2024leveraging, uddin2025health, hu2025natural}; and
(iii) they typically lack explicit grounding in external knowledge which can exacerbate the `black box' problem~\cite{rudin2019stop}. In contrast, each dialogue in \dataset is grounded in knowledge snippets retrieved from the WHO website. 

While these snippets come from a trusted source, \textit{the dataset has not been validated by healthcare professionals}. We therefore release it strictly as a multilingual \textit{language resource} for studying knowledge-grounded spoken dialogue. Clinical expert validation is beyond the scope of this dataset and benchmark paper. The health domain serves as a case study, selected to expose ethical challenges and to illustrate the potential for future development into deployable interventions. To this end, we also release a prototype system to encourage follow-up work with healthcare professionals and local communities.

\vspace{0.5mm}
\noindent \textbf{P3. Dialogues from Speakers with Diverse Backgrounds.}
Most existing dialogue datasets often fail to represent varieties \textit{within} a language, such as regional accents and dialects, due to the absence of speech data~\cite{joshi2025natural,liu2025mdseval}. \dataset addresses this limitation by providing spoken utterances recorded by native speakers sampled to reflect a diverse range of language varieties (see Figure~\ref{fig:language_diversity}).
In addition, \dataset provides demographic and sociolinguistic annotations for the speakers of each dialogue. These annotations enable systematic benchmarking of model performance beyond speech recognition and across demographic and sociolinguistic groups (e.g., knowledge retrieval accuracy by age group). While previous speech datasets such as Switchboard~\cite{godfrey1992switchboard} and Common Voice~\cite{ardila2019common} also include speaker metadata, they are not explicitly linked to dialogue tasks.

\vspace{0.5mm}
\noindent \textbf{P4. Dialogues with Coherence and Multi-Parallelism.}
Most parallel dialogue datasets are constructed via translation from an English source dataset. While this translation-based approach is cost-efficient and can natively yield parallel data across languages, it can also result in undesired `translationese' effects~\cite{Artetxe:2020emnlp}, which can reduce dialogue naturalness and inflate performance for non-English languages. To bypass the translation-based approach, \dataset adopts a \textit{bottom-up, outline-based} data collection approach~\cite{Majewska:2023cod}, which discerns between language-agnostic abstract dialogue schemata and language-specific surface realisations of the schemata (i.e, the actual utterances). As a result, \dataset provides dialogues that are both coherent and multi-parallel.

\section{\dataset}
\label{sec:dataset}

\dataset contains information-seeking dialogues in the health domain across four languages: Arabic (\ara; Afro-Asiatic), Chinese (\zho; Sino-Tibetan), English (\eng; Indo-European), and Spanish (\spa; Indo-European). The dataset comprises a total of 6,000 dialogues (1,500 per language), with 41,988 dialogue turns. As a spoken dialogue dataset, \dataset provides approximately 163 hours of user speech, recorded by native speakers representing diverse language varieties, and 208 hours of machine-generated system speech. Each dialogue turn is explicitly annotated with knowledge snippets crawled from the WHO website. In total, \dataset includes 12,045 unique snippets, of which 6,472 (4$\times$1,618) are fully parallel.

\begin{figure*}[t!]
    \centering
    \includegraphics[width=\linewidth]{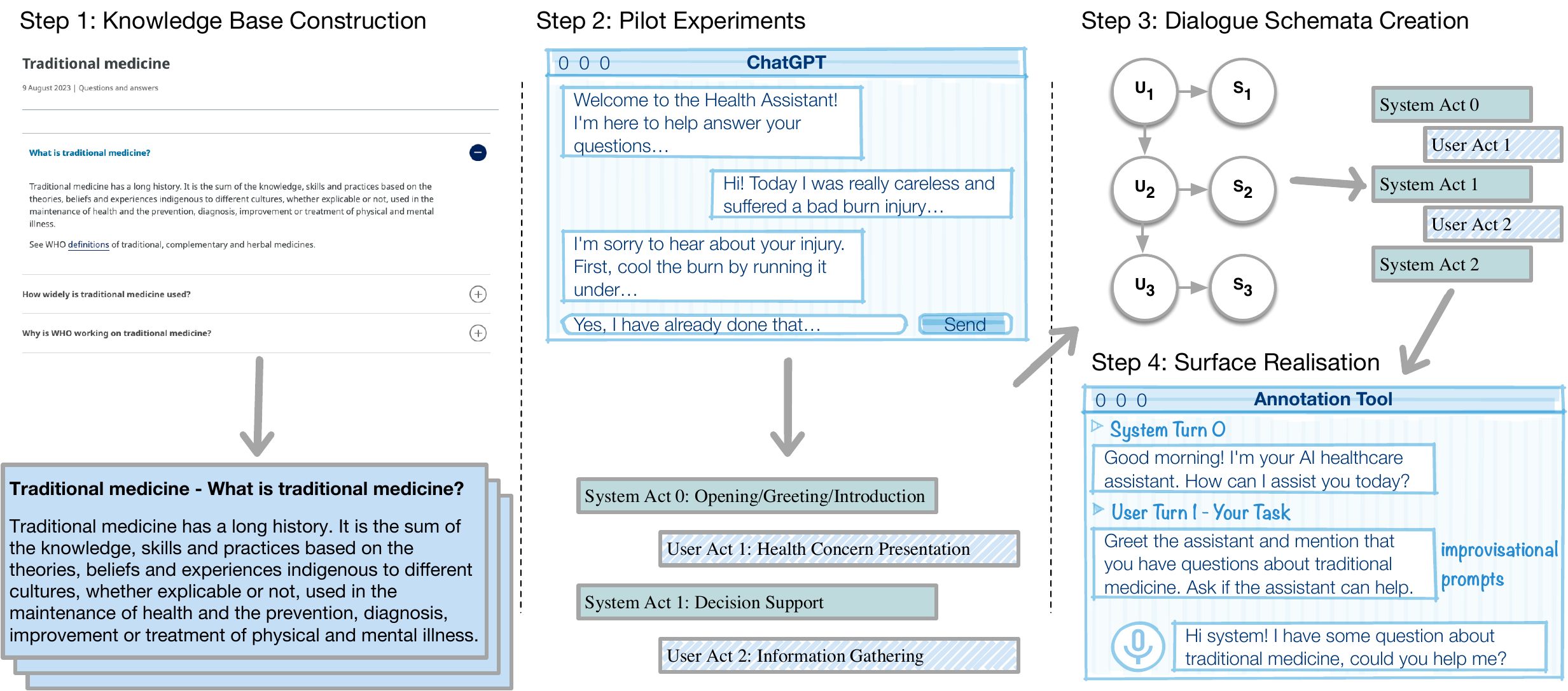}
       \caption{Overview of the data collection pipeline.
       The process consists of four main steps: 
    (i) \textit{knowledge base construction}, where we crawl knowledge snippets from the WHO website;
    (ii) \textit{pilot experiments}, where we collect 20 dialogues between a user and \texttt{gpt-4o} across hypothetical health scenarios (e.g., burns, mental health), and apply discourse analysis to identify 11 core dialogue acts;
    (iii) \textit{dialogue schemata construction}, where we model transitions between dialogue turns using a Markov chain and sample 1,500 dialogue schemata (Step 3, right). Each schema, combined with sampled knowledge snippets, is used to prompt an LLM to generate a hypothetical dialogue;
    and (iv) \textit{surface realisation}, in which annotators use LLM-generated improvisational prompts (derived from the hypothetical dialogue) to construct naturalistic spoken dialogues. User utterances are then recorded and transcribed.}

    \label{fig:pipeline}

\end{figure*}

In what follows, we describe its creation, as depicted in Figure~\ref{fig:pipeline}. Our approach involves four key steps:
\begin{enumerate*}[label=(\roman*)]
\item \textit{knowledge base construction}, in which we define the scope of the dataset and provide explicit grounding for dialogue turns;
\item \textit{pilot experiments}, where we collect a small set of 20 dialogues and analyse their high-level discourse structure;
\item \textit{dialogue schemata construction}, where we sample abstract dialogue schemata to guide LLMs in generating diverse hypothetical English dialogues;
\item \textit{surface realisation}, in which native speakers of each target language transform improvisational prompts, which are derived from the hypothetical English dialogues, into fully naturalistic dialogue turns using an outline-based approach.
\end{enumerate*}

\rparagraph{Preliminaries and Task Definition} We define the systems built upon \dataset as a conversational interface to a knowledge base, thereby specifying a bounded scope of knowledge the system is expected to use. In this setting, any generated content that cannot be verified against the knowledge base is considered as an instance of extrinsic hallucination~\cite{ji2023survey} and is discouraged.

The dataset $\mathbbm{D}$ comprises four multi-parallel sets of dialogues, denoted as $\mathbbm{D}^{\ara}$, $\mathbbm{D}^{\zho}$, $\mathbbm{D}^{\eng}$, and $\mathbbm{D}^{\spa}$, each grounded in a corresponding set of knowledge snippets, $\mathbbm{K}^{\ara}$, $\mathbbm{K}^{\zho}$, $\mathbbm{K}^{\eng}$, and $\mathbbm{K}^{\spa}$, respectively. Each knowledge snippet $\mathbf{k} \in \mathbbm{K}$ is a tuple comprising a topic, title, and content:
$\mathbf{k} = (\mathbf{topic}, \mathbf{title}, \mathbf{content})$.
Each dialogue $\mathcal{D} \in \mathbbm{D}$ is represented as a sequence of alternating user and system turns, beginning with a system introduction:
$\mathcal{D} = [\mathbf{s}_0, (\mathbf{u}_1, \mathbf{s}_1, \mathcal{K}_1, \mathbf{r}_1), \ldots, (\mathbf{u}_n, \mathbf{s}_n, \mathcal{K}_n, \mathbf{r}_n)]$,
where $\mathbf{u}_i$ and $\mathbf{s}_i$ denote the natural language utterances from the user and system, respectively; $\mathcal{K}_i \subseteq \mathbbm{K}$ is the set of knowledge snippets supporting the system response $\mathbf{s}_i$ (which may be empty); and $\mathbf{r}_i \in \{0, 1\}$ is a binary indicator specifying whether the system performs knowledge retrieval (i.e., accesses external information from $\mathbbm{K}$) in response to the user query $\mathbf{u}_i$.
Based on $\mathbf{r}_i$ and $\mathcal{K}_i$, we distinguish three scenarios:
(i) $\mathbf{r}_i = 0$ (no retrieval required, e.g., ``Hello, nice to meet you.'');  
(ii) $\mathbf{r}_i = 1$ and $|\mathcal{K}_i| \geq 1$ (retrieval with grounding);  
(iii) $\mathbf{r}_i = 1$ and $|\mathcal{K}_i| = 0$ (retrieval attempted but no supporting snippet found). Case (iii) is referred to as \textit{Out-of-Knowledge} (OOK), indicating a query that is not covered by $\mathbbm{K}$.
The initial system utterance $\mathbf{s}_0$ serves to inform the user that they are interacting with an AI system. In addition, each utterance is represented in two modalities: the audio form, denoted by $\mathbf{u}_i^{(a)}$ and $\mathbf{s}_i^{(a)}$, and the corresponding textual transcription, denoted by $\mathbf{u}_i^{(t)}$ and $\mathbf{s}_i^{(t)}$.

\rparagraph{Knowledge Base Construction}
\dataset provides a total of 12,045 knowledge snippets sourced from the WHO Questions and Answers and Fact Sheets. Specifically, the dataset includes 2,317 snippets for $\mathbbm{K}^{\ara}$, 2,431 for $\mathbbm{K}^{\zho}$, 4,785 for $\mathbbm{K}^{\eng}$, and 2,512 for $\mathbbm{K}^{\spa}$. An example snippet is shown in Figure~\ref{fig:pipeline}, where each entry consists of a $\mathbf{topic}$ (e.g., \textit{traditional medicine}), a $\mathbf{title}$ (e.g., \textit{What is traditional medicine?}), and the actual $\mathbf{content}$. Among these, 1,618 snippets per language are aligned in parallel across all four languages. We assign each aligned set a unique parallel identifier, allowing the same snippet to be indexed consistently across languages. The detailed procedure for collecting and aligning these snippets is provided in Appendix~\ref{sec:snippets_construction_appendix}. 

The topic labels are automatically derived from the inherent structure of the WHO webpages (e.g., page structure and hierarchical organisation), rather than from manual annotations. This design allows the pipeline to leverage such existing structures in other knowledge sources and does not depend on WHO-specific annotations. In cases where explicit topic metadata is not available, similarity-based clustering can be used to approximate topic groupings.

\rparagraph{Pilot Experiments}
To better understand the structure of human–machine dialogues and minimise arbitrariness in design, we conducted a pilot study by collecting 20 health consultation dialogues between 10 human users and a prototype dialogue system developed using \texttt{gpt-4o}.\footnote{For brevity, we refer to language models by short names. Full model checkpoint names are listed in Table~\ref{tab:input_lm_appendix}.} 
The system was assigned the role of a health advisor via prompting and provided health advice by leveraging its unbounded parametric knowledge. Each user was given a hypothetical health scenario, such as burns or mental disorders, and was instructed to seek health advice from the system via text. One scenario, the exact model prompt, and hyperparameters used for generation are detailed in Appendix~\ref{sec:pilot_appendix}.
These dialogues were then manually analysed using discourse analysis. Specifically, we applied Dialogue Act Theory~\citep{stolcke2000dialogue, core1997coding} to construct a dialogue act schema with 11 dialogue acts (see the full list in Example~\ref{example:dialogue_acts} in the Appendix), each representing a specific function in the dialogue, such as \textit{Information Gathering} or \textit{Care Planning and Guidance}.

\rparagraph{Dialogue Schemata Creation}
To address ethical concerns associated with collecting personal health information, we prompt \texttt{gpt-4o} to generate hypothetical dialogues. Annotators then construct human-created dialogues from these hypothetical ones. However, generating dialogues directly from LLMs without structured priors often results in repetitive or unnatural interactions~\cite{chu2024exploring, duan-etal-2024-botchat, liu-etal-2024-toad, liu-etal-2024-unlocking}. Therefore, we condition the dialogue generation process on a dialogue schema sampled from a first-order Markov chain, constructed from dialogue structures observed in our pilot experiments. The sampled dialogue act sequence (schema) serves as a high-level prompt to encourage structurally diverse dialogue trajectories, acting as a heuristic prior rather than a strict statistical constraint. Specifically, the transition between user turns is represented by a transition probability, such that each user act is conditioned on the user act from the previous turn, and each system act is conditioned solely on the current user act. Model parameters are provided in Figure~\ref{fig:transition_matrix_appendix} in the Appendix.

Finally, for each sampled dialogue schema $\mathcal{A}$, we prompt \texttt{gpt-4o} to generate a natural language dialogue $\mathcal{D}^{\eng}$ in English, conditioned on $\mathcal{A}$ and a set of sampled knowledge snippets $\mathcal{K} \subseteq \mathbbm{K}^{\eng}$. The set $\mathcal{K}$ is selected such that all $\mathbf{k} \in \mathcal{K}$ share the same $\mathbf{topic}$. To enable consistent comparisons across languages, we restrict sampling to the 1,618 parallel knowledge snippets. In addition, we apply a post-hoc modification to 10\% of the English dialogues ($\mathbbm{D}^{\eng}$) by introducing an OOK user turn: a question that cannot be answered using  $\mathbbm{K}^{\eng}$.

To construct OOK examples, we first provide the LLM with the full set of knowledge snippets for a given $\mathbf{topic}$ and prompt it to generate a user question that is not covered by the provided content.
We verify each generated question by retrieving the top 10 relevant snippets from $\mathbbm{K}^{\eng}$ using a BM25 retriever~\cite{robertson2009probabilistic}, and then prompt \texttt{gpt-4o} to assess whether the question can be adequately answered based on these snippets; if not, the question is accepted as OOK.
Next, we prompt \texttt{gpt-4o} to identify a suitable point in the dialogue to insert the OOK question. The corresponding user utterance is replaced with the OOK query, and the following system response is substituted with a refusal to answer (e.g., ``I cannot answer that based on the available information.''). The prompts used for dataset creation are provided in the publicly released codebase.

\rparagraph{Surface Realisation}  
Rather than tasking native speakers with reading aloud LLM-generated dialogues, we adopt an outline-based dialogue generation approach~\cite{Majewska:2023cod}, which mitigates potential artefacts introduced by both machine translation and direct LLM generation. In our annotation setup, each outline takes the form of a textual instruction, referred to as an \textit{improvisational prompt} in Figure~\ref{fig:pipeline}, that guides annotators in constructing user dialogue utterances. To create these prompts, we first use \texttt{gpt-4o} to generate one \textit{improvisational prompt} for each user utterance in every English dialogue in $\mathcal{D}^{\eng}$. These English \textit{prompts}, together with the corresponding LLM-generated dialogues, are then translated into the other three target languages also using \texttt{gpt-4o}.

To collect audio data, we developed a web-based annotation toolkit and detailed annotation guidelines. Figure~\ref{fig:annotation_tool} in the Appendix shows a screenshot of the annotation interface with the guidelines provided to annotators. During annotation, annotators are tasked with producing utterances based on a set of improvisational prompts and the surrounding dialogue context. Each utterance is first recorded as speech, and then transcribed in real time using \texttt{whisper-1} model~\cite{radford2023robust}. Annotators subsequently post-edit the automatically generated transcription.

\rparagraph{Duration, Cost, Annotators, and Quality Control}
While the overall project spanned more than 12 months, the actual data collection process took place over a 3-month period beginning in January 2025. The total cost of data collection was approximately \$16,000, evenly distributed across the four target languages. All annotators were native speakers of each target language, primarily consisting of professional translators recruited via \href{https://www.proz.com}{\nolinkurl{proz.com}} and university students. The released dataset includes contributions from 24 native Arabic speakers, 23 native Chinese speakers, 23 native English speakers, and 23 native Spanish speakers.

We implemented multiple quality control measures throughout the annotation process. First, annotators were required to complete a qualification round to ensure their understanding of the task; submissions were reviewed by the research team before annotators were permitted to contribute to the released dataset. Second, our web-based annotation platform included real-time validation checks that provided immediate feedback and flagged potential issues. Finally, we conducted post-collection validation: the research team manually reviewed 10\% of the dataset. For Arabic and Spanish, which were beyond the language expertise of the research team, the dialogues were translated into English for verification. Dialogues from two annotators were removed or recollected. The most frequent issues arose from annotators failing to correct ASR transcription errors. These occurred primarily in dialectal speech, where ASR systems were less robust, increasing annotator workload.

\rparagraph{Ethical and Responsible Data Creation and Use}  
This project prioritises ethical and responsible practices in both data creation and use, following the principles outlined by~\citet{rogers-etal-2021-just-think}. The study received ethics approval from the University of Cambridge, and we outline key ethical considerations below.

\rrparagraph{Terms of Use}  
Text and code are released under the MIT License. Audio data are released under a customised data use agreement that restricts use to non-commercial purposes and explicitly prohibits misuse such as voice cloning or attempts to re-identify annotators.

\rrparagraph{Privacy}  
To comply with the EU General Data Protection Regulation (GDPR), we acted as a data controller and collected only the minimum amount of personal data required for the project. All participants provided informed consent by signing a \textit{Participant Consent Form} prior to data collection. The dataset consists entirely of hypothetical dialogues with predefined content, thereby minimising the risk of unintentionally including personal data.

\rrparagraph{Compensation}  
Annotators were compensated \$200 for contributing 75 dialogues, corresponding to an approximate hourly rate of \$20.

\rparagraph{Data Structure and Statistics}
All dialogues in \dataset consist of parallel utterances in four languages.  Figure~\ref{fig:example_dialogue} in the Appendix shows an example of multi-parallel dialogues. For each user turn, we provide a spoken utterance recorded by a native speaker, its corresponding transcription, and an LLM-generated alternative for comparison. Each system turn includes an LLM-generated response and corresponding machine-generated audio, annotated with the supporting set of knowledge snippets and a knowledge retrieval indicator flag. Textual data is released in \texttt{JSON} format, while audio data is provided as mono-channel, 16-bit WAV files sampled at 16\,kHz.

\begin{figure}[!t]
    \centering
    \includegraphics[width=\linewidth]{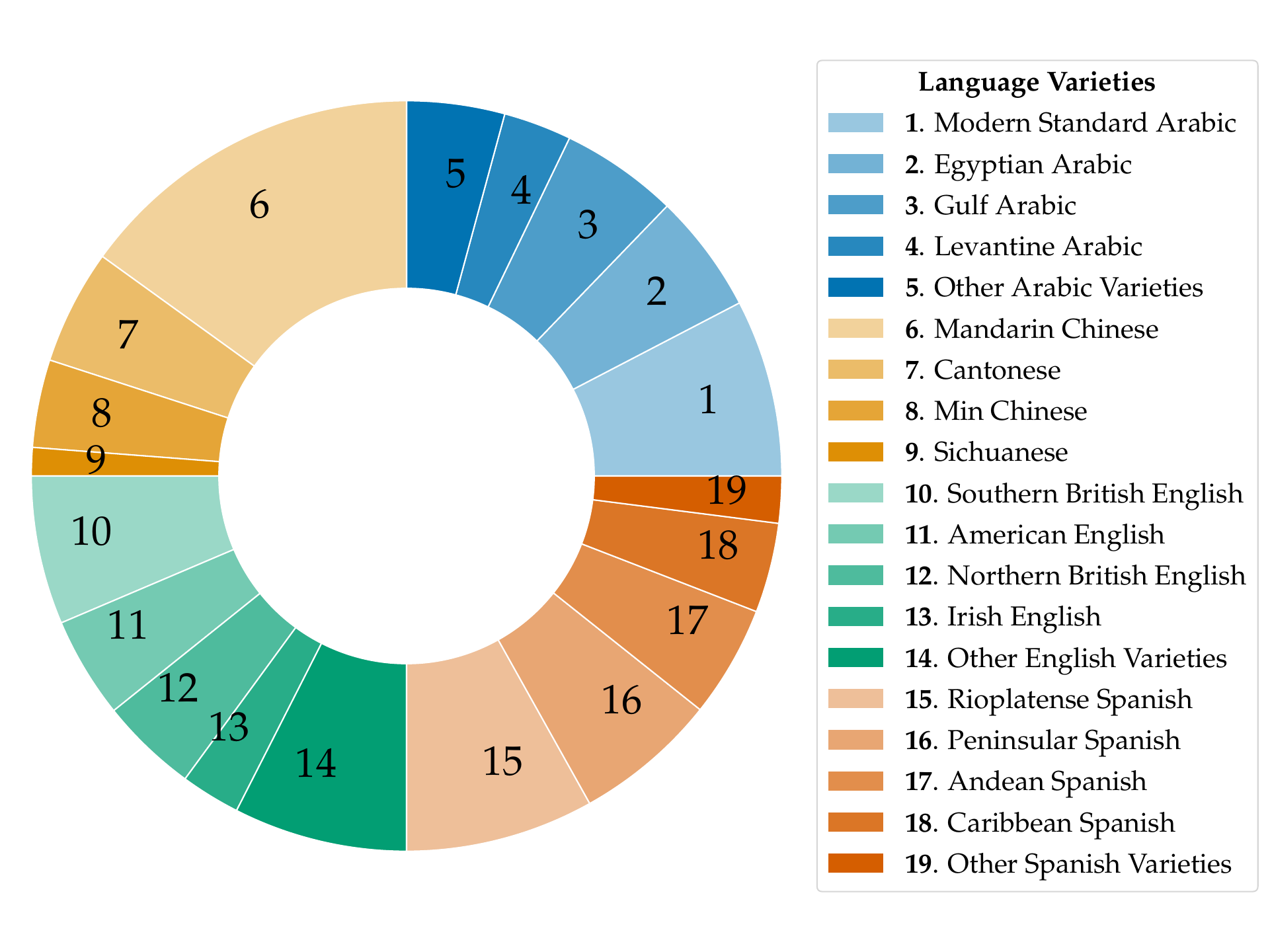}
    \caption{Distribution of dialogues across the top four language varieties for each language. Less represented varieties are grouped into the \textit{Others} category.}
    \label{fig:language_diversity}

\end{figure}

\dataset includes spoken dialogues spanning a wide range of language varieties for each target language, as shown in Figure~\ref{fig:language_diversity}. This linguistic diversity is complemented by a balanced gender distribution and a broad age range among annotators (see Figures~\ref{fig:annotator_gender} and~\ref{fig:annotator_age} in the Appendix). As shown in Table~\ref{tab:token_stats} in the Appendix, human-authored user utterances are consistently longer and more lexically diverse than those generated by LLMs. In English, for instance, human utterances contain on average 35.71 tokens, compared to 18.66 tokens for LLM-generated counterparts, with a substantially larger vocabulary size. Similar trends are observed across Arabic, Chinese, and Spanish.

\section{\dataset as a Benchmark}
\label{sec:experiments}

\dataset serves as a multilingual benchmark for evaluating LLMs and other system components within a retrieval-augmented generation (RAG) pipeline~\cite{lewis2020retrieval, asai2023self}. In this section, we outline the overall system architecture, define each component task, and present initial benchmark results for these tasks.

\rparagraph{System Pipeline}  
The dialogue system defined in \dataset takes as input a dialogue history up to time step $t$, represented as:
$\mathcal{H}_t = [\mathbf{s}_0, (\mathbf{u}_1, \mathbf{s}_1), \ldots, (\mathbf{u}_{t-1}, \mathbf{s}_{t-1}), \mathbf{u}_t^{(a)}]$, where $\mathbf{u}_t^{(a)}$ denotes the user query at time step $t$ in audio form.
The system is tasked to perform the following sequence of operations:

\rrparagraph{ASR} 
The ASR model maps $\mathbf{u}_t^{(a)}$ to its transcription $\mathbf{u}_t^{(t)}$. The updated dialogue history is then used by downstream modules: $\mathcal{H}_t = [\mathbf{s}_0, (\mathbf{u}_1, \mathbf{s}_1), \ldots, (\mathbf{u}_{t-1}, \mathbf{s}_{t-1}), \mathbf{u}_t^{(t)}]$.

\rrparagraph{Retrieval Turn Classification}  
The system predicts whether the current user query requires external knowledge: $\hat{\mathbf{r}}_t = f_{\text{classification}}(\mathcal{H}_t)$, where $\hat{\mathbf{r}}_t \in \{0, 1\}$ is a binary variable indicating whether the response should be grounded in external knowledge ($\hat{\mathbf{r}}_t = 1$) or can be generated from context alone ($\hat{\mathbf{r}}_t = 0$).

\rrparagraph{Knowledge Selection} If $\hat{\mathbf{r}}_{t}=1$, the system proceeds in two stages: (i) a high-recall
retrieval model produces a fixed-size candidate set: $\mathcal{K}_{t}^{\mathrm{cand}} =f_{\mathrm{retrieve}}\!\bigl(\mathcal{H}_{t},\mathbbm{K}\bigr)$; (ii) a high-precision filtering model that produces the final support: $\hat{\mathcal{K}}_{t} =f_{\mathrm{filter}}\!\bigl(\mathcal{H}_{t}, \mathcal{K}_{t}^{\mathrm{cand}}\bigr)$, giving the final support set
$\hat{\mathcal{K}}_{t}\subseteq\mathcal{K}_{t}^{\mathrm{cand}}
\subseteq\mathbbm{K}$.

\rrparagraph{Response Generation}  
The system generates a response $\hat{\mathbf{s}}_t^{(t)} = f_{\text{generation}}(\mathcal{H}_t)$ if $\hat{\mathbf{r}}_t = 0$, or $\hat{\mathbf{s}}_t^{(t)} = f_{\text{generation}}(\mathcal{H}_t, \hat{\mathcal{K}}_t)$ if $\hat{\mathbf{r}}_t = 1$. When $\hat{\mathbf{r}}_t = 1$ but no relevant snippets are retrieved (i.e., $\hat{\mathcal{K}}_t = \emptyset$), the system is required to explicitly indicate that the query is OOK.

\rrparagraph{TTS}  
The TTS model converts the textual response $\hat{\mathbf{s}}_t^{(t)}$ to its spoken form $\hat{\mathbf{s}}_t^{(a)}$.

\begin{table*}[t!]
\centering
\def\arraystretch{0.9}
{\scriptsize
\resizebox{\textwidth}{!}{%
\begin{tabular}{l cc c cc c c c cc c cc}
\toprule
\multirow{2}{*}{\textbf{Language}} 
& \multicolumn{2}{c}{\textbf{ASR}} & 
& \multicolumn{2}{c}{\textbf{TTS}} & 
& \textbf{Turn Cls.} & 
& \multicolumn{2}{c}{\textbf{Knowledge Retrieval}} & 
& \multicolumn{2}{c}{\textbf{Knowledge Filtering}$^{*}$} \\
\cmidrule(lr){2-3} \cmidrule(lr){5-6} \cmidrule(lr){8-8} \cmidrule(lr){10-11} \cmidrule(lr){13-14}
& \textbf{WER} $\downarrow$ & \textbf{CER} $\downarrow$ 
& & \textbf{MCD} $\downarrow$ & \textbf{CER} $\downarrow$
& & \textbf{Acc.} $\uparrow$ 
& & \textbf{R@10 (T)} $\uparrow$ & \textbf{R@10 (S)} $\uparrow$
& & \textbf{EM} $\uparrow$ & \textbf{OOK Recall} $\uparrow$ \\
\midrule
Arabic  & 0.23 & 0.07 & & 12.08 & 0.10 & & 95.39 & & 65.88 & 0.20 & & 34.27 & 0.00 \\
Chinese & 0.24 & 0.14 & & 11.46 & 0.17 & & 95.23 & & 70.63 & 0.23 & & 39.19 & 14.29 \\
English & 0.03 & 0.01 & & 11.44 & 0.06 & & 96.30 & & 75.72 & 0.52 & & 44.29 & 42.86 \\
Spanish & 0.02 & 0.01 & & 10.84 & 0.07 & & 95.93 & & 71.82 & 0.42 & & 39.54 & 14.29 \\
Average & 0.13 & 0.06 & & 11.46 & 0.10 & & 95.71 & & 71.01 & 0.34 & & 39.32 & 17.36 \\
\bottomrule
\end{tabular}
}
}
\caption{Performance of the best-performing model for each component task in \dataset. The best model for ASR is \texttt{whisper-1}; for TTS, \texttt{gpt-4o-mini-tts}; for turn classification (Turn Cls.), \texttt{XLM-R\textsubscript{large}}; for text-to-text knowledge retrieval (evaluated using R@5 (T)), \texttt{text-embedding-3L}; for speech-to-text knowledge retrieval (evaluated using R@5 (S)), \texttt{CLAP}; and for knowledge filtering, \texttt{gpt-4.1}. ($^*$) For knowledge filtering, performance is reported on a randomly sampled 10\% subset of the test set. }
\label{tab:main_results}

\end{table*}

This pipelined design is susceptible to error propagation, particularly from the ASR model. While each component could in principle operate directly on speech input, current speech-native models are not yet robust enough for multi-turn dialogue across languages (see supporting evidence in Table~\ref{tab:retrieval_appendix}). We therefore adopt a pipelined implementation in this paper and release \dataset to enable future research.

\rparagraph{ASR and TTS}
We evaluate a set of ASR models: \texttt{whisper-1} and \texttt{phi-4-MM-Inst}~\cite{abouelenin2025phi}, as well as TTS using \texttt{gpt-4o-mini-tts}~\cite{achiam2023gpt}. For TTS, we condition generation on speaker demographic variables, including age group, primary language, place of origin, region of residence, and education level.
Table~\ref{tab:main_results} presents evaluation results for the best-performing ASR and TTS models. Additional results for the remaining models are provided in Table~\ref{tab:asr_tts_appendix} in the Appendix. ASR models are evaluated using Word Error Rate (WER) and Character Error Rate (CER), while TTS models are assessed using Mel Cepstral Distortion (MCD)~\cite{407206} and CER measured via ASR.

\rparagraph{Retrieval Turn Classification}
We evaluate \texttt{XLM-R\textsubscript{large}}~\cite{conneau-etal-2020-unsupervised} fine-tuned on a training set of 500 dialogues and \texttt{LLaMA3.1-8B-Inst}~\cite{grattafiori2024llama} with 10 randomly sampled in-context examples from the same set. Table~\ref{tab:main_results} reports the best-performing model, while Table~\ref{tab:knowledge_selection_appendix} provides the full results. Both models achieve over 90\% accuracy, indicating the simplicity of the task, since 75.5\% of dialogue turns require knowledge retrieval.

\rparagraph{Knowledge Selection}
In standard RAG pipelines, the retrieval model returns a candidate set of knowledge snippets, and the language model implicitly performs knowledge filtering during generation by attending to relevant content through its internal attention mechanism~\cite{lewis2020retrieval}. However, in high-stakes domains such as healthcare, interpretability is essential. To this end, we explicitly model the knowledge filtering process, following prior benchmarks that treat knowledge selection as a standalone task~\cite{dinan2018wizard}. Accordingly, we establish three benchmarks: (i) multilingual text-to-text retrieval, (ii) multilingual speech-to-text retrieval, and (iii) explicit knowledge filtering.

We evaluate text-to-text retrieval using a set of text encoders: \texttt{text-embedding-3L}, \texttt{gte-multilingual-B}~\cite{zhang2024mgte}, \texttt{MiniLM-L12-v2}~\cite{reimers-gurevych-2019-sentence}, \texttt{NV-Embed-v2}~\cite{lee2025nvembed} and the statistical method BM25. We also evaluate speech-to-text retrieval using a set of multimodal multilingual encoders, including \texttt{CLAP}~\cite{10095969} and \texttt{SpeechT5}~\cite{ao-etal-2022-speecht5}. All models are evaluated on a parallel subset of knowledge snippets. Table~\ref{tab:main_results} reports the best-performing model for each retrieval task, while Table~\ref{tab:retrieval_appendix} in the Appendix provides detailed results across all models. Models are evaluated using recall, precision, F1 score, and Maximal Marginal Relevance (MMR).

Based on the full results in Table~\ref{tab:retrieval_appendix}, we observe that larger and more recent encoders, such as \texttt{text-embedding-3L}, consistently outperform smaller models like \texttt{MiniLM-L12-v2}. Additionally, multilingual text encoders exhibit notable performance disparities across languages: English achieves the highest retrieval scores, while Arabic performs the lowest, with a gap of nearly 10 points top-5 recall (R@5). Finally, we find that all evaluated multimodal encoders perform near random chance on the speech-to-text retrieval task, highlighting the difficulty of this setting and the need for future research on cross-modal alignment.

For knowledge filtering, we evaluate two approaches:
(i) a threshold-based method, which retains snippets whose retrieval scores, measured by cosine similarity between the dialogue history and each candidate snippet, exceed a fixed threshold; and
(ii) LLM-based methods, in which a language model is prompted to assess the relevance of each candidate snippet and retain only those it predicts relevant. We evaluate \texttt{gpt-4.1-nano} and \texttt{LLaMA3.1-8B-Inst} on the full test set, and evaluate other models from the OpenAI GPT family on 10\% of the test set due to the high cost of running multilingual experiments at scale. Full results are presented in Table~\ref{tab:knowledge_selection_appendix} (Appendix). Each model receives as input the top 5 retrieved snippets, selected using \texttt{text-embedding-3L}, and is evaluated using the Exact Match (EM) score against the ground-truth snippet set. Also, we report OOK Recall, which measures whether the model correctly returns an empty set when the query cannot be answered based on the knowledge base.

\begin{figure}[!t]
    \centering
    \includegraphics[width=\linewidth]{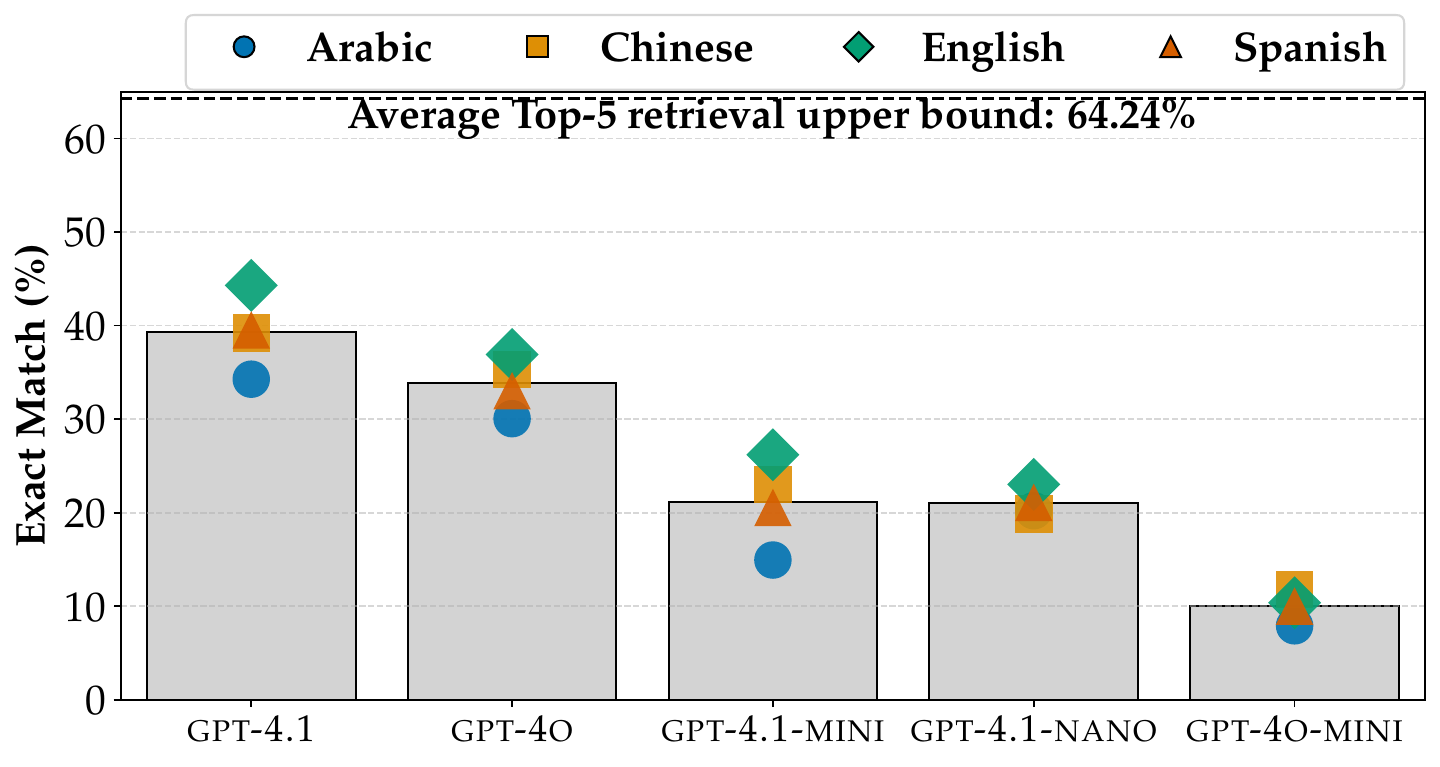}
    \caption{
    Knowledge filtering accuracy measured by Exact Match score for OpenAI models on 10\% of the test set. Each model is tasked with selecting relevant knowledge snippets from the top 5 candidates, ranked using \texttt{text-embedding-3L}.}
    \label{fig:openai_knowledge_filtering}

\end{figure}

Our results highlight consistent performance disparities across languages in the knowledge selection pipeline. Despite the fully parallel experimental setup, English consistently achieves the highest retrieval and filtering accuracy, while Arabic shows the lowest performance across all models. Furthermore, as shown in Table~\ref{tab:selection_analysis}, increasing the number of retrieved candidates does not necessarily improve accuracy. While a larger candidate set raises the likelihood that the correct snippet is recalled, it also introduces more distracting snippets, which lowers filtering accuracy. This finding highlights that simply extending the input context is insufficient, and that a well-designed retrieval-augmented pipeline remains essential. In addition, Figure~\ref{fig:openai_knowledge_filtering} shows that larger and more capable LLMs achieve higher performance on the deductive reasoning task of knowledge filtering, also leaving substantial room for improvement. This further demonstrates that \dataset can serve as a multilingual benchmark for evaluating the deductive reasoning capabilities of LLMs.

The observed cross-lingual performance disparities are consistent with patterns reported in prior multilingual literature. Recent meta-analyses of multilingual benchmarks~\cite{hu-etal-2025-quantifying} report similar language rankings, with English and Spanish typically outperforming Arabic and Chinese. Comparable trends have also been documented in multilingual ASR studies~\cite{Pratap2020MLSAL, yadav-sitaram-2022-survey}. Importantly, we observe consistent disparities across multiple components of the pipeline, suggesting a systematic pattern rather than a task-specific artefact.

\begin{table}[t!]
\centering
\def\arraystretch{0.9}
{\scriptsize
\resizebox{\columnwidth}{!}{%
\begin{tabular}{l cccc}
\toprule
\multirow{2}{*}{\textbf{Language}} 
& \multirow{2}{*}{\textbf{Threshold} } 
& \multicolumn{3}{c}{\textbf{LLM-based}} \\
 \cmidrule(lr){3-5}
& 
& \footnotesize Top-5
& \footnotesize Top-10
& \footnotesize Top-50 \\
\midrule
Arabic  & 6.26 & 19.96 & 12.58 & 10.85 \\
Chinese & 6.61 & 19.86 & 17.15 & 12.28 \\
English & 6.88 & 23.02 & 23.33 & 18.72 \\
Spanish & 6.46 & 21.09 & 19.55 & 11.03 \\
Average & 6.55 & 21.05 & 18.15 & 13.72 \\
\bottomrule
\end{tabular}
}
}
\caption{Knowledge filtering accuracy measured by Exact Match score. \textbf{Threshold} refers to a fixed similarity score used to retain relevant snippets, tuned on a validation set. \textbf{LLM@Top-$k$} denotes filtering performed by \texttt{gpt-4.1-nano} over the top-$k$ retrieved snippets by \texttt{text-embedding-3L}. Each system turn in \dataset is supported by at most 5 ground-truth snippets.}

\label{tab:selection_analysis}
\end{table}

\begin{figure}[h]
    \centering
    \includegraphics[width=\linewidth]{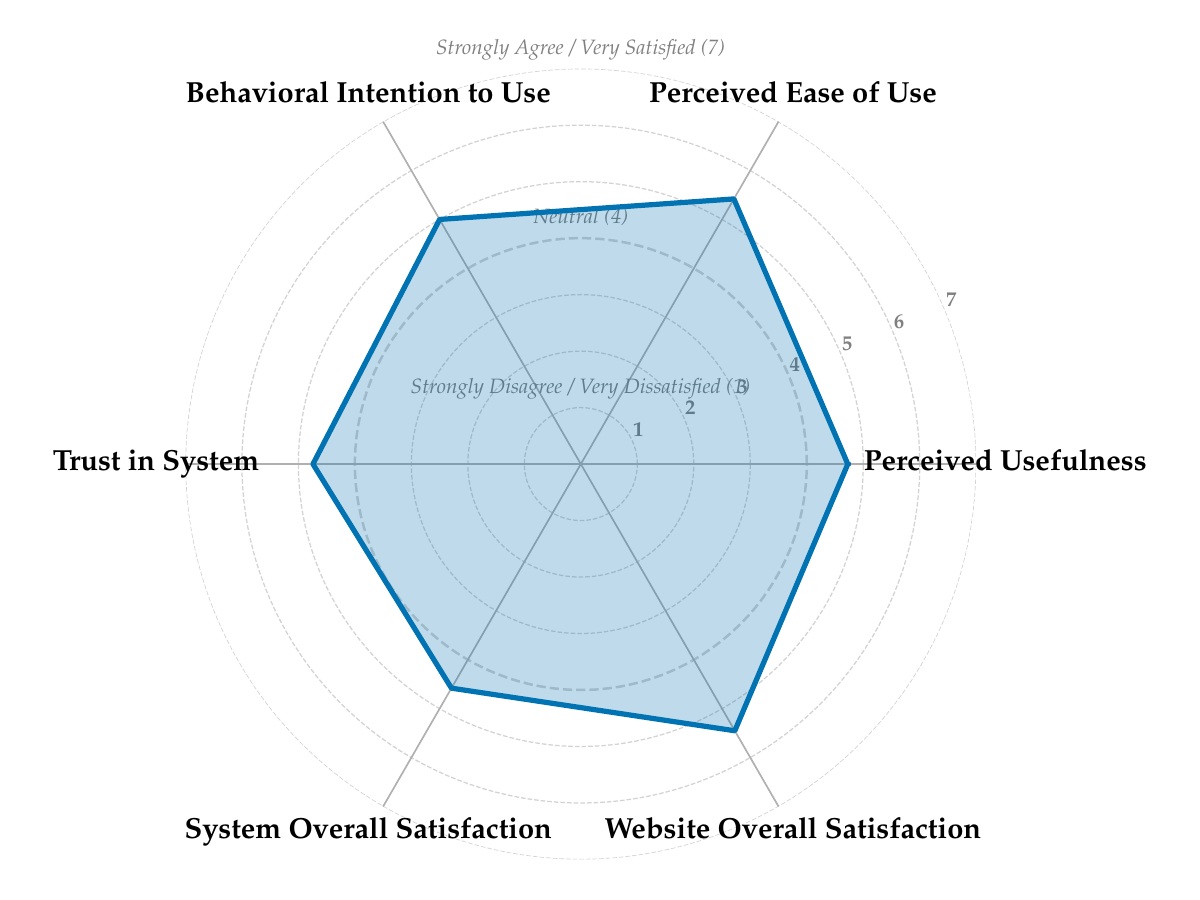}
    \caption{Average human ratings across key constructs, as reported by 25 participants. These constructs were measured using the TAM2-based questionnaire shown in Figure~\ref{fig:human_eval_tool_appendix}.}
    \label{fig:user_eval_appendix}
\end{figure}

\rparagraph{User Perceptions of Dialogue Systems}
We conducted a human evaluation experiment based on the Technology Acceptance Model 2 (TAM2) framework~\cite{venkatesh2000theoretical} to assess user acceptance of the developed dialogue system in comparison to the WHO website. A TAM2-based questionnaire was administered to 25 participants, covering key constructs such as \textit{Perceived Usefulness} and \textit{Perceived Ease of Use}. All participants were fluent English speakers and did not receive any additional training. Figure~\ref{fig:human_eval_tool_appendix} (Appendix) shows a screenshot of the evaluation interface, including the participant instructions, TAM2-based questionnaire, and our prototype dialogue system. Our system supports both text and speech interaction. For each system response, if available, the supporting evidence is displayed to the user.

Figure~\ref{fig:user_eval_appendix} presents the average ratings across key constructs.  The system was implemented using \texttt{gpt-4.1} as the backbone LLM. Overall, participants reported a positive attitude toward and acceptance of the system. While users found the system easy to use and generally useful, it received comparatively lower scores for perceived trustworthiness. Furthermore, the system's overall satisfaction rating is lower than that of the WHO website. Qualitative feedback highlights the need for improvements in system output quality, user trust, and more proactive dialogue policies.

This TAM2-based evaluation is intended as an illustrative demonstration of how dialogue systems built on \dataset can be evaluated in a standardised and reproducible manner, rather than as a comprehensive assessment of cross-lingual usability or trustworthiness. Large-scale cross-lingual human evaluation remains future work.

\section{Conclusion}
\label{sec:conclusion}
We present a large-scale data collection process that produces a multilingual, multi-parallel spoken dialogue dataset for benchmarking multilingual dialogue systems. The dataset provides 6,000 dialogues and 163 hours of user speech, recorded by native speakers representing diverse language varieties across Arabic, Chinese, English, and Spanish. This dataset addresses a critical resource gap for benchmarking multilingual spoken dialogue systems and enables future evaluation of speech-native models. We benchmark a range of NLP tasks, including ASR, TTS, text-to-text and speech-to-text retrieval, and the deductive reasoning task of knowledge filtering, establishing baselines for future research.

Enabled by this dataset, future research can conduct controlled analyses of performance disparities, not only across languages, but also within language varieties and user demographics such as gender and age, and propose methods to mitigate them. Beyond standard NLP benchmarking, we release not only the dataset but also a prototype dialogue system and a complete toolkit for data collection and system evaluation. These resources will enable the research community to collect similar datasets at scale, develop spoken dialogue systems, and evaluate their performance with real-world users.

\section*{Limitations}

While \dataset provides, to our knowledge, the first large-scale multilingual spoken dialogue benchmark, it is important to acknowledge its limitations.

\rparagraph{Data Collection with Synthetic Dialogues and Outline-Based Generation}
The content of \dataset was generated with LLMs and has not been validated by healthcare professionals. We therefore release it strictly as a multilingual \textit{language resource} for studying knowledge-grounded spoken dialogue. Although such validation could in principle be added as an extension of our pipeline, it would require substantial resources and collaboration with medical experts, which lies beyond the scope of this work. Our contribution is instead methodological: a reproducible pipeline for constructing multilingual, multi-parallel spoken dialogue datasets at scale. By combining LLMs with human annotations, we reduce both cost and privacy risks, and release a benchmark dataset that enables experiments previously not feasible with existing resources.

\rparagraph{Real-World Needs and Cultural Nuances}
Dialogue systems should remain responsive to real-world needs, which evolve both temporally and geographically (e.g., during the COVID-19 pandemic or in regions with a high prevalence of non-communicable diseases). Continuously retraining LLMs to reflect such changes is often infeasible~\cite{lewis2020retrieval, clusmann2023future}, motivating our use of RAG-based design. While grounding in WHO snippets ensures full parallelism across languages, it inevitably limits cultural adaptation, since the materials are not tailored to local practices. Addressing this would require collaboration with healthcare and cultural experts, which falls beyond the scope of this study. Nevertheless, by releasing a parallel benchmark, we provide a foundation for future work on culturally adapted dialogue systems.

\rparagraph{Cross-Lingual and Cross-Study Evaluations}
Our benchmark results reveal consistent disparities across languages, aligning with prior findings in multilingual NLP~\cite{pmlr-v119-hu20b, hu-etal-2023-systematic, xuan2025mmlu}. However, the relative ranking of languages varies across tasks and benchmarks, reflecting a broader challenge: multilingual evaluations are often confounded by target language choice, task design, and model selection. Thus, most benchmarks converge only on the qualitative conclusion that high-resource languages consistently outperform others. Recent work has attempted to quantify such disparities more systematically~\cite{hu2025quantifying}, but a comprehensive cross-study comparison remains an open challenge.

\rparagraph{End-to-End Speech-Based Evaluation}
While \dataset is designed to support fully speech-based system evaluation, our benchmark study follows a pipelined architecture that decomposes the system into ASR, retrieval, generation, and TTS components. This design choice reflects the current state of the field: existing speech-native language models and multimodal encoders are not yet sufficiently robust to support end-to-end spoken dialogue benchmarking at scale. In practice, the extremely low performance of current models in fully speech-based settings limits the interpretability of quantitative comparisons at present. As speech-native language models mature, \dataset can support future evaluations of fully speech-based dialogue systems.

\section*{Acknowledgements}
This work is supported by the Cambridge–LMU Strategic Partnership grant.
This work is also supported by the UK Research and Innovation (UKRI) Frontier Research Grant EP/Y031350/1 EQUATE awarded to Anna Korhonen.
Songbo Hu is supported by the Cambridge International Scholarship. 
Ivan Vuli\'{c} is supported by a Royal Society University Research Fellowship, \textit{`Inclusive and Sustainable Language Technology for a Truly Multilingual World'} (no 221137).

\bibliography{custom}

\appendix

\section{Supplementary Details for Replication}
\label{appendix:supplementary_detials}

We provide supplementary information to support the replication of the data creation process and benchmark experiments described in this paper. We disclose the use of AI assistants for code writing and editorial assistance during the preparation of this work.

\begin{figure}[h]
    \centering
    \includegraphics[width=\linewidth]{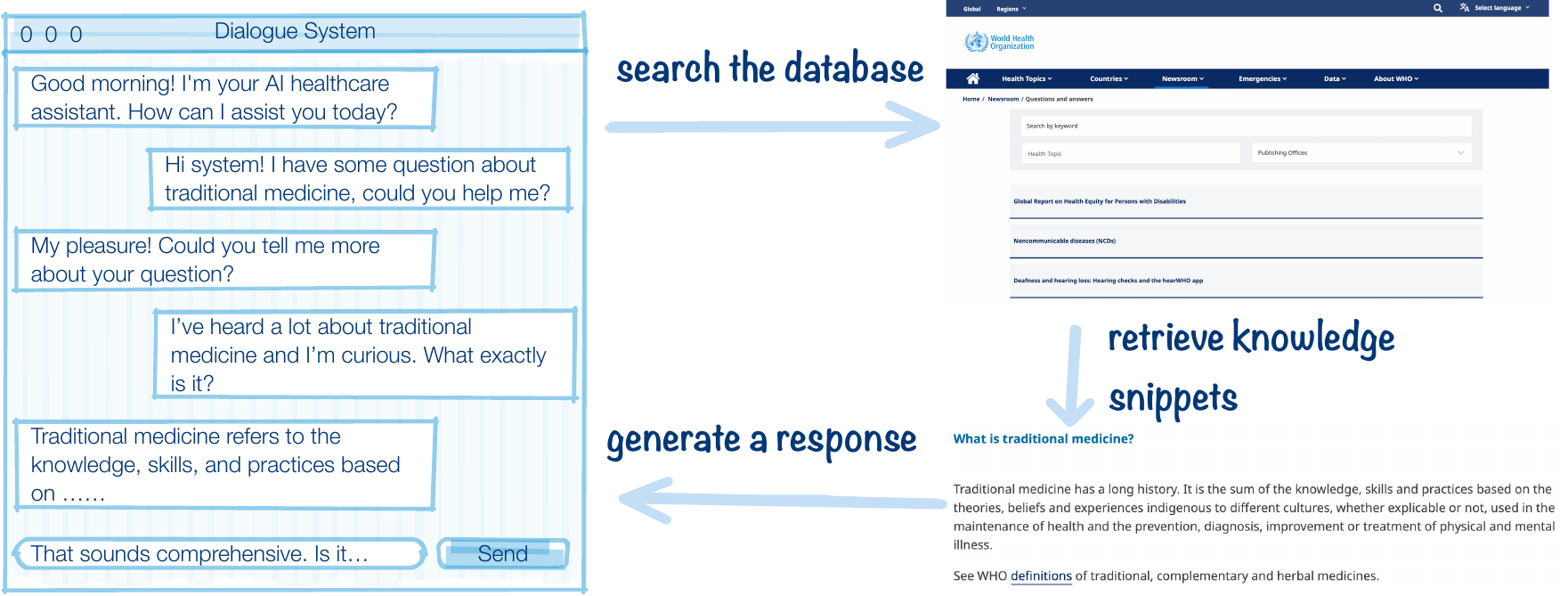}
    \caption{An illustration of dialogue systems based on \dataset. Each dialogue turn is explicitly grounded in external knowledge snippets sourced from the WHO website. While the illustration shows a text-based dialogue in English, \dataset supports both text and speech in Arabic, Chinese, English, and Spanish.}
    \label{fig:exmaple_task}
\end{figure}

\vspace{0.5mm}
\noindent
\textbf{Figure~\ref{fig:exmaple_task}} illustrates the intended use case of dialogue systems developed using \dataset, highlighting the integration of grounding in external health knowledge.

\subsection{Knowledge Base Construction}
\label{sec:snippets_construction_appendix}

\textbf{Example~\ref{example:json-snippet}} shows a sample knowledge snippet collected in English in \dataset.

\begin{lstlisting}[language=json, caption={Example of a health knowledge snippet in JSON format.},
    label={example:json-snippet}]
{
  "url": "https://www.who.int/news-room/questions-and-answers/item/traditional-medicine",
  "language": "ENG",
  "data": {
    "type": "qa_pair",
    "topic": "Traditional medicine",
    "title": "What is traditional medicine?",
    "content": "Traditional medicine has a long history. It is the sum of the knowledge, skills and practices based on the theories, beliefs and experiences indigenous to different cultures, whether explicable or not, used in the maintenance of health and the prevention, diagnosis, improvement or treatment of physical and mental illness."
  },
  "parallel_data": true,
  "parallel_identifier": "questions-and-answers/item/traditional-medicine::0",
  "unique_identifier": "6a85e2b5-ee53-493f-82ca-26488110b593"
}
\end{lstlisting}

\begin{figure}[h]
    \centering
    \includegraphics[width=\linewidth]{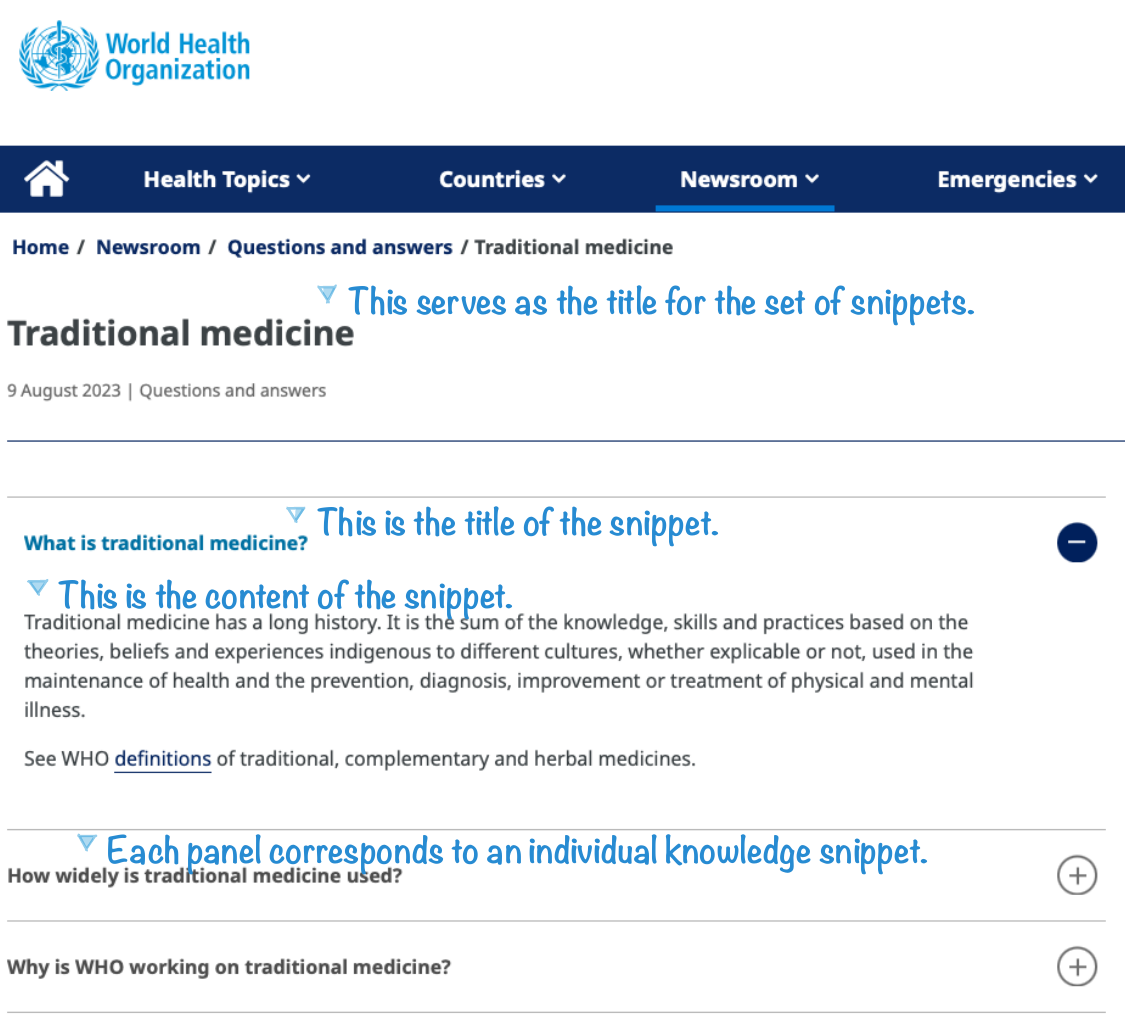}
    \caption{A screenshot of a WHO webpage. The figure is annotated to show how each component of the webpage corresponds to the attributes of a knowledge snippet.}
    \label{fig:exmaple_snippet}
\end{figure}

\vspace{0.5mm}
\noindent \textbf{Knowledge Snippet Collection.} We crawled webpages from the WHO website on 11 May 2025, as described in \S\ref{sec:dataset}. As shown in Figure~\ref{fig:exmaple_snippet}, each collapsible panel corresponds to a single knowledge snippet, with the page heading serving as the $\mathbf{topic}$, the panel title as the $\mathbf{title}$, and the panel content as the $\mathbf{content}$ of the snippet. Multiple snippets can be derived from the same page, all sharing the same $\mathbf{topic}$.
Overall, we extracted 4,785 English snippets ($\mathbbm{K}^{\eng}$) from 590 webpages, 2,317 Arabic snippets ($\mathbbm{K}^{\ara}$) from 299 webpages, 2,431 Chinese snippets ($\mathbbm{K}^{\zho}$) from 303 webpages, and 2,512 Spanish snippets ($\mathbbm{K}^{\spa}$) from 307 webpages.

\vspace{0.5mm}
\noindent \textbf{Knowledge Snippet Alignment.}  
Each WHO health topic page typically contains multiple knowledge snippets, which may not always appear in the same order or have exact one-to-one correspondence across different languages. In order to construct four parallel sets of knowledge snippets, we identify the largest possible set of matched snippets that are aligned across all four languages for each page. We approach this problem by assuming that the English knowledge snippets form a superset of the others; that is, $\mathbbm{K}^{\ara}, \mathbbm{K}^{\zho}, \mathbbm{K}^{\spa} \subseteq \mathbbm{K}^{\eng}$. The task is then transformed into assigning each snippet in the other three languages to a corresponding snippet in English. We define the alignment as a set of functions that map each non-English snippet to a corresponding English snippet: $f^{\lan} : \mathbbm{K}^{\lan} \rightarrow \mathbbm{K}^{\eng} \quad \text{for } \lan \in \{\ara, \zho, \spa\}$, where \( f^{\lan}(\mathbf{k}) \) returns the English snippet in \( \mathbbm{K}^{\eng} \) that is semantically equivalent to snippet \( \mathbf{k} \in \mathbbm{K}^{\lan} \).
This alignment problem can be modelled as a \textit{linear sum assignment problem}~\citep{martello1987linear}, which seeks an optimal assignment of ‘tasks’ to ‘workers’ that minimises the total cost. In our case, the cost is defined as the semantic distance between snippets in English and those in the other three languages. We compute pairwise semantic distance as \( 1 - \text{cosine similarity} \) between snippet embeddings, where each snippet is represented by the concatenation of its $\mathbf{title}$ and $\mathbf{content}$. Embeddings are generated using OpenAI’s \texttt{text-embedding-3-L} model. The optimal assignments are computed using the \textit{Hungarian algorithm}~\citep{martello1987linear}.

\subsection{Pilot Experiments}
\label{sec:pilot_appendix}

\textbf{Example~\ref{example:burn_scenario}} shows one example scenario used in the pilot experiment. It was generated by ChatGPT and verified by the research team.

\begin{examplebox}[example:burn_scenario]{Burn injury scenario}
You recently experienced a burn injury while cooking at home. The affected area is painful and appears red and swollen. You're unsure about the severity of the burn and whether you should seek medical attention. Questions arise about how to properly care for the burn at home, what signs indicate a need for professional medical help, and how long it will take to heal. These concerns are important to address promptly to ensure proper treatment and avoid complications.
\end{examplebox}

\vspace{0.5mm}
\noindent
\textbf{Example~\ref{example:pilot_system_prompt}} shows the system prompt used to develop the pilot system. The model checkpoint is \texttt{gpt-4o}, with a temperature of 0.5 and a top\_p of 0.9 used for the sampling method.

\begin{examplebox}[example:pilot_system_prompt]{Pilot system prompt}
        You are a health advisor and please try to answer the following question from a patient. 
        You should provide a brief response to the patient's question.
        Your response should also be coherent with the dialog history.
        Users might not always have access to medical professionals. Please try your best to answer their questions
        Please only output the response but nothing else.
\end{examplebox}

\vspace{0.5mm}
\noindent
\textbf{Example~\ref{example:dialogue_acts}} presents the 11 dialogue acts we constructed, each with an example utterance, based on the 20 pilot dialogues described in \S\ref{sec:dataset}.

\begin{examplebox}[example:dialogue_acts]{Dialogue Acts in \dataset}
\vspace{3pt}

    \textbf{1. Opening:} The \textit{system} initiates the conversation with a greeting and an introduction to its role or the service provided.\\
    \textit{Example (System):} Hello, I'm your virtual health assistant. How can I help you today?

    \textbf{2. Health Concern Presentation:} The \textit{user} states their primary health concern, symptom, or question.\\
    \textit{Example (User):} Hey, I burned my hand cooking last week. It's really painful, red, and swollen.

    \textbf{3. Information Gathering:} The \textit{system} asks clarification questions to gather more context about the user's symptoms or medical history.\\
    \textit{Example (System):} Were you vaccinated for yellow fever before your trip?

    \textbf{4. Explanation / Medical Education:} The \textit{system} provides in-depth information or educates the user about their condition, treatment options, and preventive measures.\\
    \textit{Example (System):} If the burn is larger than 3 inches or on your face, hands, or joints, you should definitely see a doctor.

    \textbf{5. Care Planning and Guidance:} The \textit{system} offers specific advice on managing the health issue, including treatment options, preventive measures, lifestyle modifications, and self-care techniques.\\
    \textit{Example (System):} Until you see a doctor, keep the burn clean and covered with a sterile, non-stick bandage.

    \textbf{6. Decision Support:} The \textit{user or system} may discuss different options, relevant risks and benefits, and explore user preferences.\\
    \textit{Example (System):} It's important to consider your options and what feels right for you. You can also seek support from a trusted friend, family member, or a professional counsellor.

    \textbf{7. Healthcare System Navigation:} The \textit{user or system} may discuss guidance on navigating the healthcare system, including finding a provider, making an appointment, and understanding insurance coverage and costs.\\
    \textit{Example (System):} You can find a local urgent care centre or call your primary care doctor to schedule an appointment.

    \textbf{8. Legal and Ethical Considerations:} The \textit{user or system} may discuss legal and ethical considerations, including informed consent and patient rights.\\
    \textit{Example (System):} In the UK, your medical records are confidential and protected by law.

    \textbf{9. Privacy and Confidentiality:} The \textit{user or system} may inquire about, or proactively assure, the privacy and confidentiality of the user's information.\\
    \textit{Example (System):} Your information is safe with us. We take your privacy very seriously.

    \textbf{10. Emotional Support:} The \textit{system} offers emotional support, empathy, and reassurance to the user.
    \textit{Example (System):} I'm sorry to hear that you're going through this. It's completely normal to feel scared and overwhelmed.

    \textbf{11. Closing:} The \textit{system} ends the conversation with a summary, an offer of further assistance, or a farewell.
    \textit{Example (System):} You're welcome! Take care, and I hope you feel better soon. Goodbye!

\vspace{3pt}
\label{example3}
\end{examplebox}

\subsection{Dialogue Schemata Creation}
\label{sec:schemata_construction_appendix}

\begin{figure}[h]
    \centering
    \includegraphics[width=\linewidth]{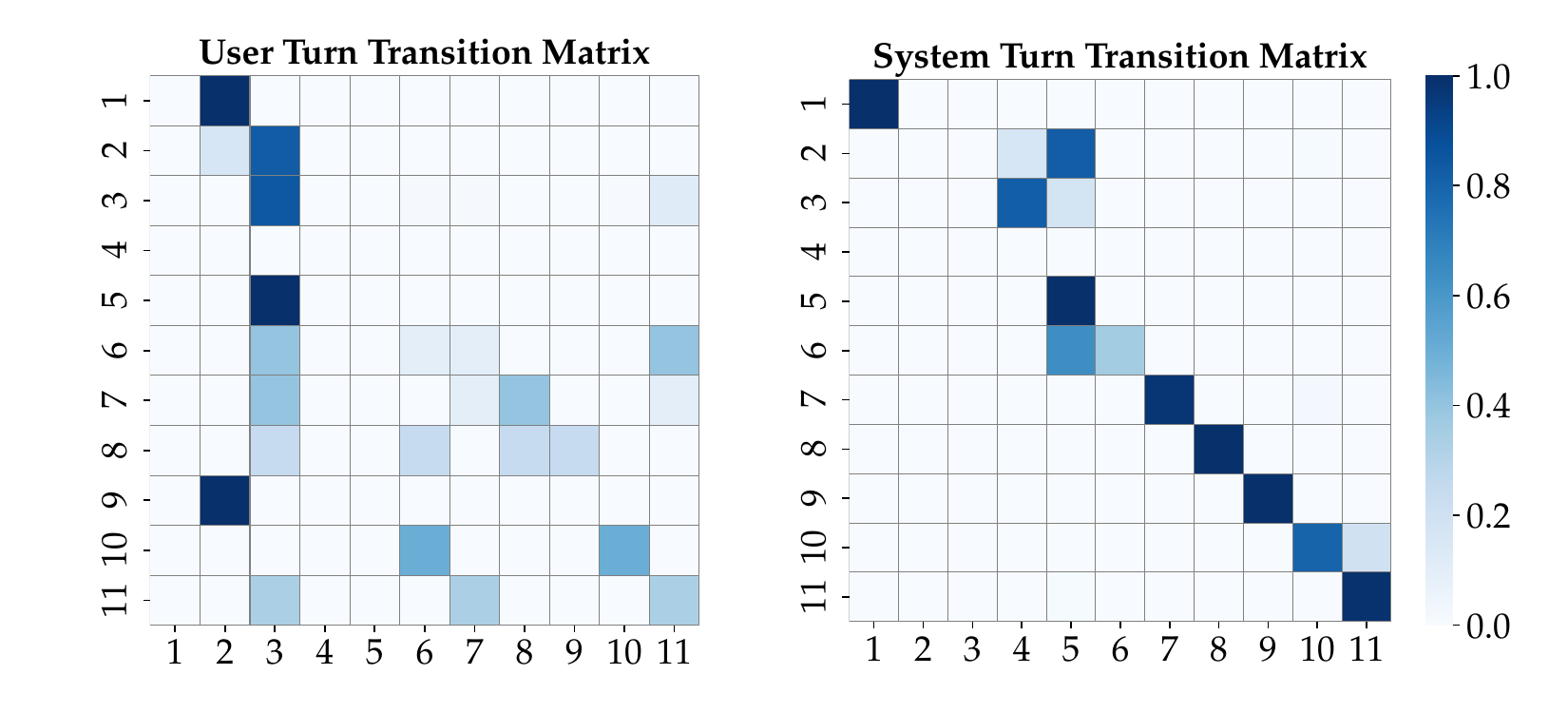}
\caption{Transition probabilities in our Markov model. The left plot shows user turn transitions, namely $P(\mathbf{a}_i^u \mid \mathbf{a}_{i-1}^u)$, while the right plot shows system turn transitions, $P(\mathbf{a}_i^s \mid \mathbf{a}_i^u)$. Role indices correspond to the discourse role schema presented in Example~\ref{example:dialogue_acts}.}
    \label{fig:transition_matrix_appendix}
\end{figure}

\textbf{Figure~\ref{fig:transition_matrix_appendix}} shows the transition probabilities in our hierarchical Markov model. Let $\mathbf{a}_i^u$ and $\mathbf{a}_i^s$ denote the discourse acts associated with the $i$-th user and system turns, respectively, and let $\mathbf{a}_0^s$ denote the initial system act, which is fixed across all dialogues (i.e., an \textit{Opening} act introducing the system). The full \textit{dialogue schema} is defined as: $\mathcal{A} = [\mathbf{a}_0^s, \mathbf{a}_1^u, \mathbf{a}_1^s, \ldots, \mathbf{a}_n^u, \mathbf{a}_n^s]$. We factorise the probability of the schema (excluding the fixed first act) as: $P(\mathcal{A}) = \prod_{i=1}^{n} P(\mathbf{a}_i^u \mid \mathbf{a}_{i-1}^u) \cdot P(\mathbf{a}_i^s \mid \mathbf{a}_i^u)$, where $P(\mathbf{a}_i^u \mid \mathbf{a}_{i-1}^u)$ represents the user-to-user transition probabilities, and $P(\mathbf{a}_i^s \mid \mathbf{a}_i^u)$ models the system’s response act conditioned on the current user act.

\subsection{Surface Realisation.}
\label{sec:surface_realisation_appendix}

\begin{figure*}[h]
    \centering
    \includegraphics[width=0.75\linewidth]{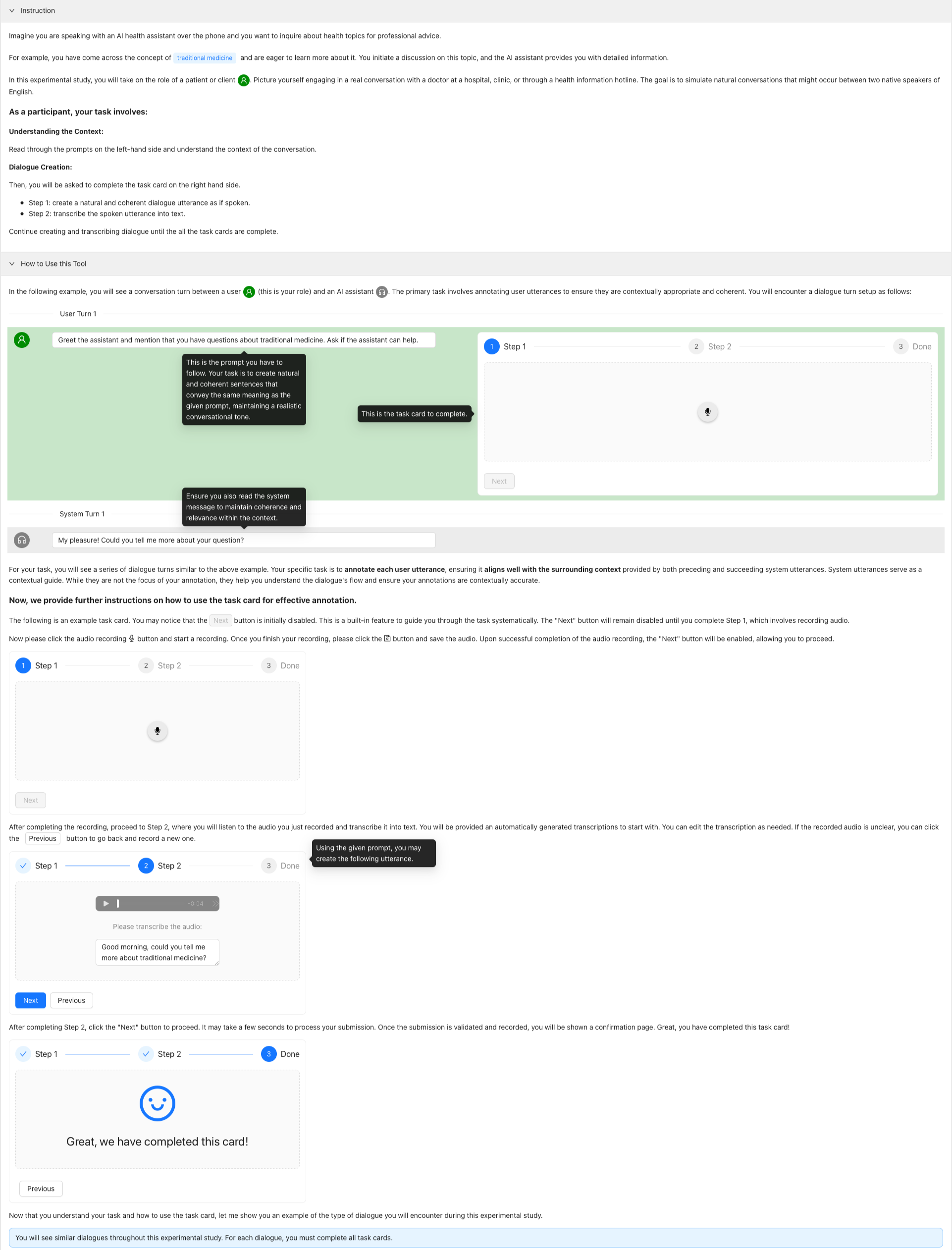}
    \caption{Screenshot of the annotation interface with the guidelines shown to annotators during English data collection.}
    \label{fig:annotation_tool}
\end{figure*}

\textbf{Figure~\ref{fig:annotation_tool}} shows a screenshot of the annotation interface with the guidelines shown to annotators. Annotators for Arabic, English, and Spanish were presented with guidelines in English, except for Chinese annotators, who received the guidelines in Chinese. This exception was made possible due to the availability of native Chinese-speaking researchers on our team who translated the website.

\subsection{Examples and Statistics of \dataset}
\label{sec:dataset_statistics_appendix}

\begin{figure*}[t!]
    \centering
    \includegraphics[width=\linewidth]{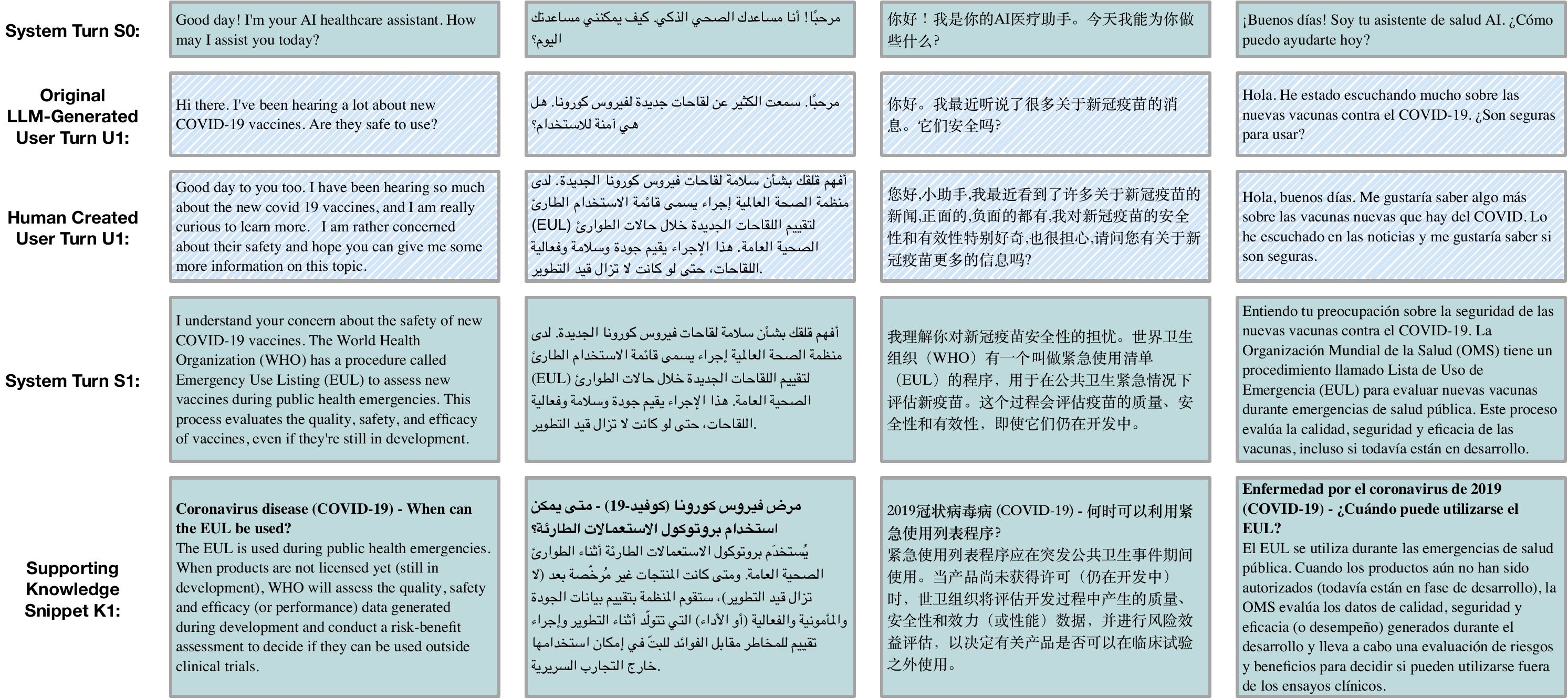}
    \caption{Example set of parallel dialogues in four languages, English, Arabic, Chinese, and Spanish, extracted from the \dataset dataset. Due to space limitations, we show only the first three turns of each dialogue. For each user turn, both the LLM-generated and the human-produced utterances are provided. As shown, human-authored utterances tend to be more complex and conversational than those generated by the LLM. The dialogue ID for this example is {\textsc{lan}\_12}.}
    \label{fig:example_dialogue}
\end{figure*}

\textbf{Figure~\ref{fig:example_dialogue}} presents a set of parallel dialogues in four languages, English, Arabic, Chinese, and Spanish, drawn from the \dataset dataset.

\begin{figure}[h]
    \centering
    \includegraphics[width=\linewidth]{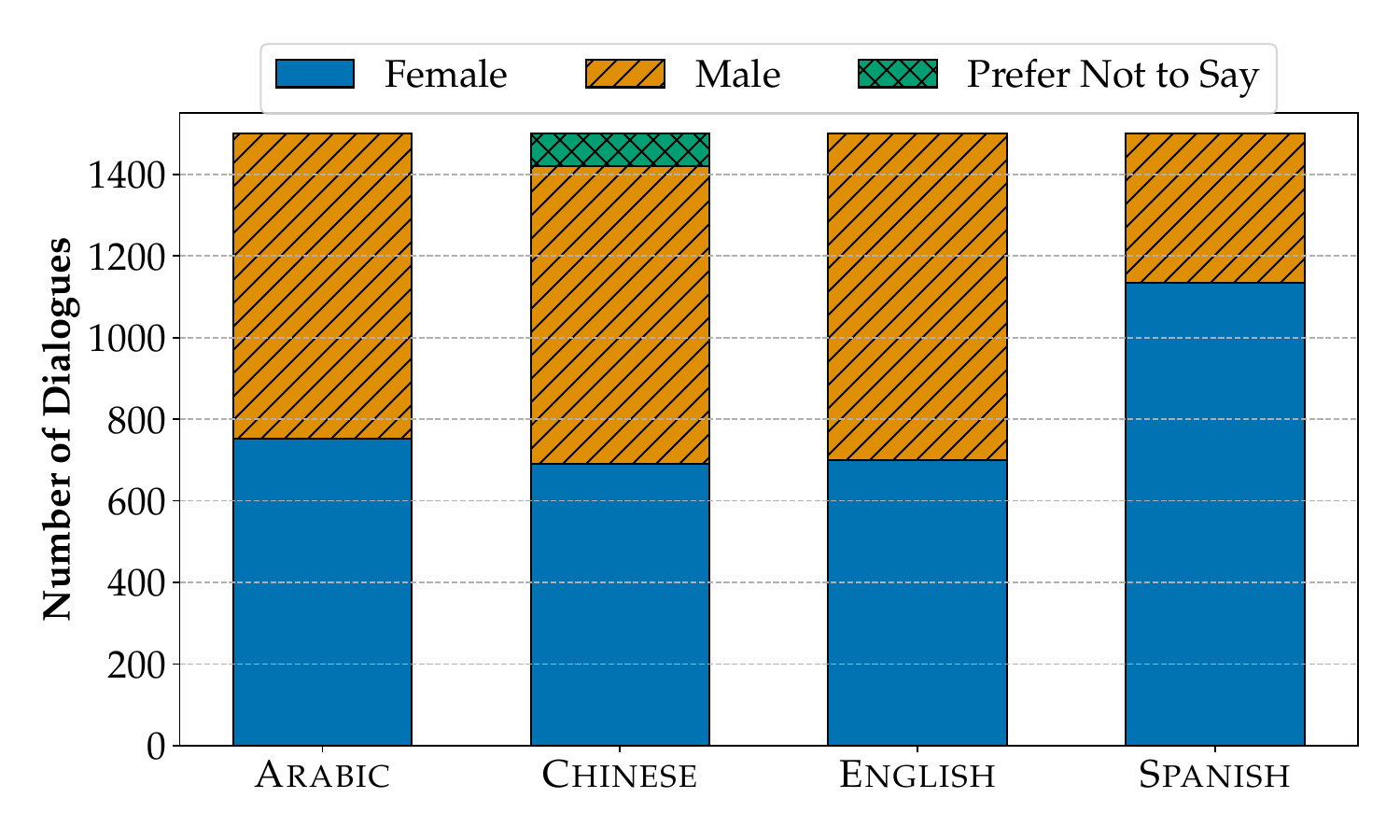}
    \caption{Distribution of dialogues by annotator gender for each language in \dataset.}
    \label{fig:annotator_gender}
\end{figure}

\begin{figure}[h]
    \centering
    \includegraphics[width=\linewidth]{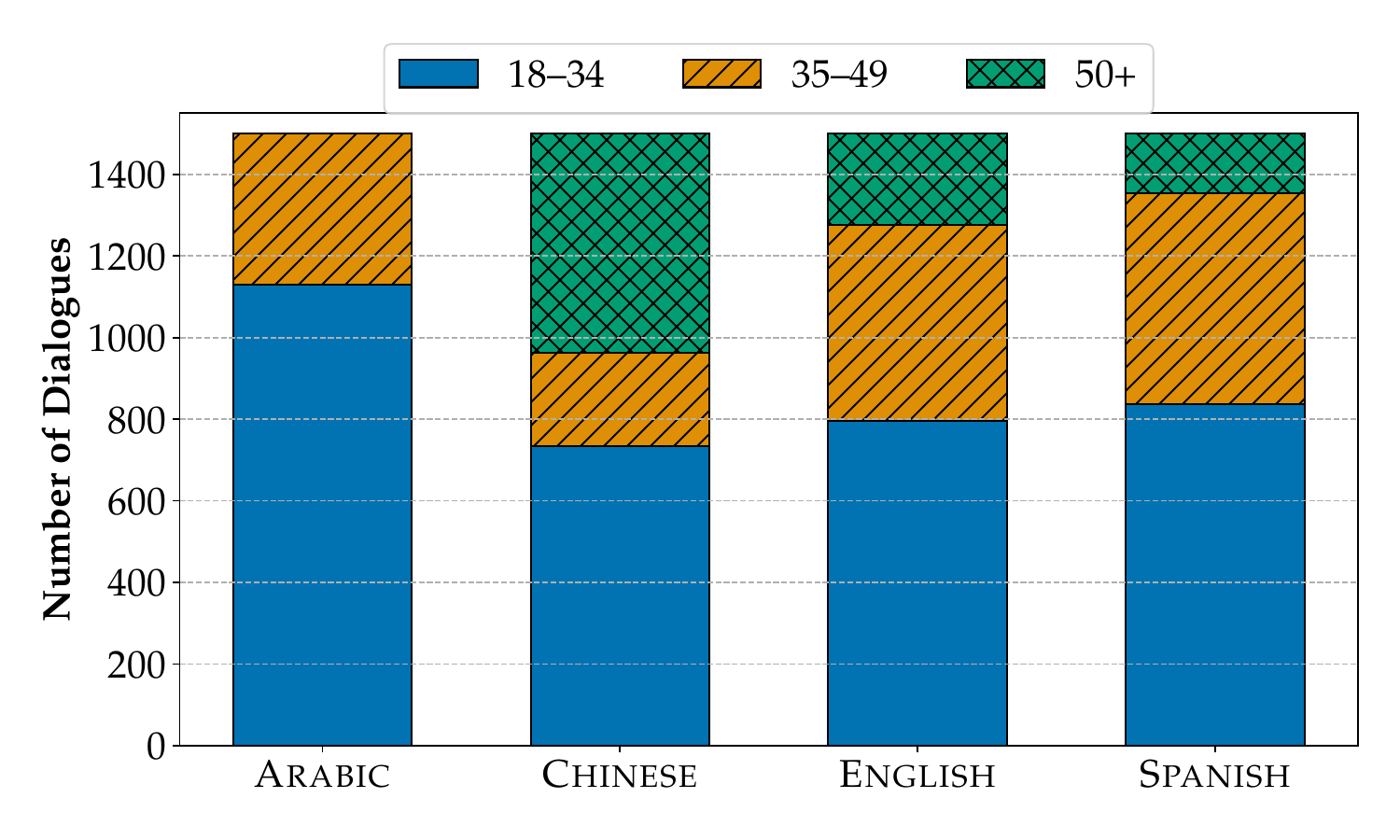}
    \caption{Distribution of dialogues by annotator age group for each language in \dataset. The original data includes finer-grained age groups, which we cluster into broader categories for visualisation.}
    \label{fig:annotator_age}
\end{figure}

\vspace{0.5mm}
\noindent
\textbf{Figure~\ref{fig:annotator_gender}} and \textbf{Figure~\ref{fig:annotator_age}} show the distribution of dialogues by annotator gender and age group, respectively, for each language.

\vspace{0.5mm}
\noindent
\textbf{Table~\ref{tab:token_stats}} reports word-level and subword-level statistics across the four languages. 
Across all languages, human-authored user utterances are consistently longer and more lexically diverse than those generated by LLMs. 
Dialogues contain an average of 6.5 user turns. 
In English, for example, human utterances contain 35.71 tokens on average, compared to 18.66 tokens for LLM-generated counterparts, with a substantially larger vocabulary size. 
Similar patterns are observed in Arabic, Chinese, and Spanish. 
These results indicate that the outline-based data collection methodology effectively elicits more diverse and naturalistic user utterances than direct LLM generation.

\begin{table*}[t!]
\centering
\def\arraystretch{0.9}
{\scriptsize
\resizebox{\textwidth}{!}{%
\begin{tabular}{l ccc ccc ccc}
\toprule
\multirow{2}{*}{\textbf{Language}} 
& \multicolumn{3}{c}{\textbf{User Utterances (Word)}} 
& \multicolumn{3}{c}{\textbf{Generated Utterances (Word)}} 
& \multicolumn{3}{c}{\textbf{User Utterances (LLaMA Subword)}} \\
\cmidrule(lr){2-4} \cmidrule(lr){5-7} \cmidrule(lr){8-10}
& \textbf{\# Tokens} & \textbf{\# Words} & \textbf{TTR} 
& \textbf{\# Tokens} & \textbf{\# Words} & \textbf{TTR} 
& \textbf{\# Tokens} & \textbf{\# Words} & \textbf{TTR} \\
\midrule
Arabic   & 356,673 & 24,165 & 0.068 & 154,832 & 7,374  & 0.048 & 695,958 & 2,477 & 0.0036 \\
English  & 348,042 & 6,327  & 0.018 & 181,838 & 3,586  & 0.020 & 355,782 & 6,846 & 0.0192 \\
Spanish  & 369,805 & 10,141 & 0.027 & 167,672 & 5,396  & 0.032 & 507,561 & 5,275 & 0.0104 \\
Chinese  & 293,954 & 11,153 & 0.038 & 165,107 & 4,161  & 0.025 & 376,343 & 3,723 & 0.0099 \\
\midrule
Average  & 342,619 & 12,947 & 0.038 & 167,862 & 5,129  & 0.031 & 483,661 & 4,580 & 0.0108 \\
\bottomrule
\end{tabular}
}
}
\caption{Comparison of word-level and subword-level statistics across four languages.
\textbf{User Utterances} correspond to human-annotated transcriptions provided by annotators, while \textbf{Generated Utterances} correspond to LLM-generated responses.  
\textbf{\# Tokens} denotes the total number of tokens across all utterances. 
\textbf{\# Words} refers to the number of unique tokens (i.e., vocabulary size). 
\textbf{TTR} represents the type-token ratio, measuring lexical diversity within each language.
Tokenisation is performed at the word level using CAMeL Tools~\cite{obeid-etal-2020-camel} for Arabic, jieba ({\href{https://github.com/fxsjy/jieba}{\nolinkurl{github.com/fxsjy/jieba}}})
 for Chinese, and NLTK~\cite{bird2006nltk} for English and Spanish, while subword-level analysis relies on the unified \texttt{LLaMA3.1-8B-Inst} tokeniser.}

\label{tab:token_stats}
\end{table*}

\subsection{Experimental Setup}
\label{sec:setup_appendix}

\textbf{Table~\ref{tab:input_lm_appendix}} lists all the language models we used in this work, along with their respective checkpoints in the Huggingface repository and the OpenAI API.

\begin{table}[h]
\centering
\def\arraystretch{0.80}
{\scriptsize
\resizebox{0.9\columnwidth}{!}{%
\begin{tabular}{l l}
\toprule
\textbf{Model} & \textbf{Checkpoint} \\ \midrule

\multicolumn{2}{c}{\cellcolor{Gray}{\textbf{Huggingface}}} \\ \midrule
\texttt{whisper-L-v3}    & openai/whisper-large-v3 \\
\texttt{phi-4-MM-Inst}    & microsoft/Phi-4-multimodal-instruct \\
\texttt{XLM-R\textsubscript{large}}  & xlm-roberta-large \\ 
\texttt{LLaMA3.1-8B-Inst}    & meta-llama/Llama-3.1-8B-Instruct \\
\texttt{gte-multilingual-B}    & Alibaba-NLP/gte-multilingual-base \\
\texttt{MiniLM-L12-v2}    & sentence-transformers/all-MiniLM-L12-v2 \\
\texttt{NV-Embed-v2}    & nvidia/NV-Embed-v2 \\
\texttt{SpeechT5}    & microsoft/speecht5\_asr \\

\midrule

\multicolumn{2}{c}{\cellcolor{Gray}{\textbf{OpenAI}}} \\ \midrule
\texttt{gpt-4.1}  & gpt-4o-2024-05-13 \\
\texttt{gpt-4.1-mini}  & gpt-4.1-mini-2025-04-14 \\
\texttt{gpt-4.1-nano}  & gpt-4.1-nano-2025-04-14 \\
\texttt{gpt-4o} (data construction)  & gpt-4o-2024-05-13 \\
\texttt{gpt-4o} (benchmark)    & gpt-4o-2024-11-20 \\
\texttt{gpt-4o-mini}    & gpt-4o-mini-2024-07-18 \\
\texttt{whisper-1}    & whisper-1 \\
\texttt{gpt-4o-mini-tts}    & gpt-4o-mini-tts \\
\texttt{text-embedding-3L}    & text-embedding-3-large \\
\bottomrule
\end{tabular}%
}
}
\caption{Language models used in our experiments, along with their corresponding HuggingFace or OpenAI checkpoints. Note that the \texttt{gpt-4o} model used for benchmarking is a more recent version than the one used during data construction. For the \texttt{CLAP} model, we use the \texttt{630k-audioset-best.pt} checkpoint from its official GitHub repository: \href{https://github.com/LAION-AI/CLAP}{\nolinkurl{github.com/LAION-AI/CLAP}}.}

\label{tab:input_lm_appendix}
\end{table}

\begin{figure*}[t!]
    \centering
    \includegraphics[width=0.7\linewidth]{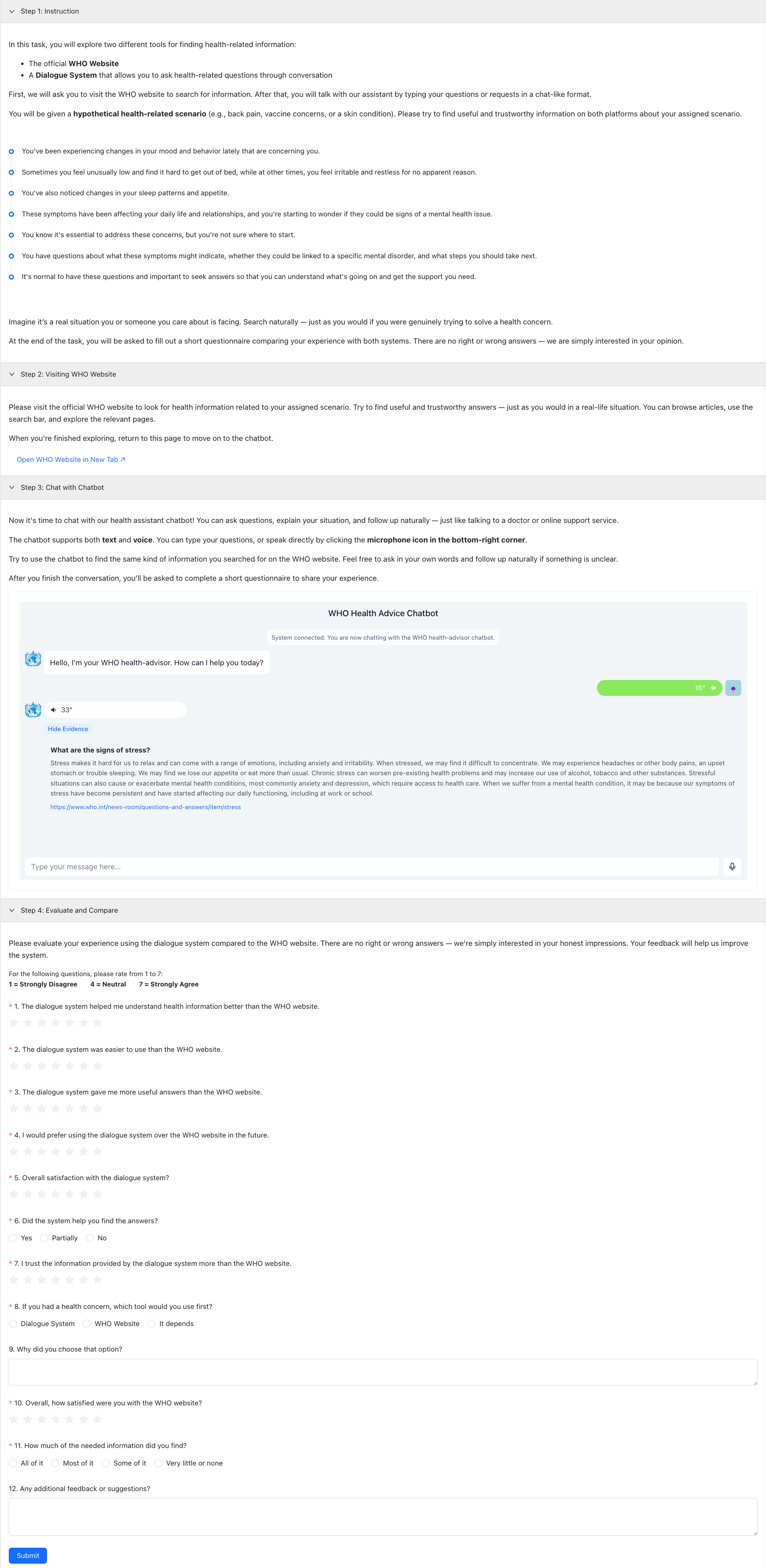}
    \caption{Screenshot of the human evaluation interface with guidelines shown to annotators. The screenshot also illustrates the user interface of our prototype dialogue system, which can be embedded within a webpage or used as a stand-alone application. The system supports both text and speech interaction. For each system response, if available, the corresponding supporting evidence can be displayed to the user.}
    \label{fig:human_eval_tool_appendix}
\end{figure*}

\vspace{0.5mm}
\noindent
\textbf{Figure~\ref{fig:human_eval_tool_appendix}} shows a screenshot of the human evaluation interface, including the guidelines provided to annotators. Questions 1 and 3 measure \textit{Perceived Usefulness}, Question 2 measures \textit{Perceived Ease of Use}, Question 4 measures \textit{Behavioural Intention to Use}, Question 7 measures \textit{Trust in System}, Question 5 measures \textit{Overall Satisfaction with the Dialogue System}, and Question 10 measures \textit{Overall Satisfaction with the WHO website}. The image also illustrates the user interface of our prototype dialogue system, which can be either embedded in a webpage or deployed as a stand-alone application. The system supports both text and speech interactions.

\clearpage

\section{Additional Results for Benchmarking}

This section presents experimental results that complement the main retrieval benchmarks discussed in \S\ref{sec:experiments}.

\vspace{0.5mm}
\noindent
\textbf{Table~\ref{tab:asr_tts_appendix}} reports the full evaluation results of the ASR and TTS models. WER assumes word-level tokenisation based on white-space, which is not directly applicable to Chinese. To address this, we pre-tokenise Chinese transcriptions using the \textit{jieba} segmentation tool.

\vspace{0.5mm}
\noindent
\textbf{Table~\ref{tab:retrieval_appendix}} shows the complete evaluation results for both text-to-text and speech-to-text retrieval tasks.

\vspace{0.5mm}
\noindent
\textbf{Table~\ref{tab:knowledge_selection_appendix}} presents evaluation results for retrieval turn classification and knowledge filtering across four languages.

\vspace{0.5mm}
\noindent
\textbf{Table~\ref{tab:response_generation_appendix}} presents the complete evaluation results for response generation, using BLEU~\cite{papineni2002bleu}, ROUGE~\cite{lin-2004-rouge}, and METEOR~\cite{banerjee2005meteor} as evaluation metrics. However, traditional reference-based metrics offer limited insight into the actual quality and utility of system responses from the perspective of end users. Moreover, a key limitation of this work is that the system responses have not been validated by healthcare professionals. As such, \dataset should not be used as a ground-truth reference for evaluating response generation models in the health domain.

\begin{table*}[!htbp]
\centering
\def\arraystretch{0.95}
{\scriptsize
\resizebox{0.8\textwidth}{!}{%
\begin{tabular}{ll ccc ccc}
\toprule
\rowcolor{Gray}
\textbf{Language} & \textbf{Model} 
& \textbf{WER} $\downarrow$ & \textbf{CER} $\downarrow$ 
& \textbf{MCD} $\downarrow$ & \textbf{CER via ASR} $\downarrow$ & \textbf{Task} \\
\cmidrule(lr){1-2} \cmidrule(lr){3-4} \cmidrule(lr){5-6} \cmidrule(lr){7-7}

\multirow{4}{*}{Arabic} 
& \texttt{whisper-1}    & 0.23 & 0.07 & ---  & ---  & ASR \\
& \texttt{phi-4}               & 5.89 & 5.79 & ---  & ---  & ASR \\
& \texttt{gpt-4o-mini-tts}     & ---  & ---  & 12.08 & 0.10 & TTS \\
\hdashline

\multirow{4}{*}{Chinese} 
& \texttt{whisper-1}    & 0.24 & 0.14 & ---  & ---  & ASR \\
& \texttt{phi-4}               & 1.03 & 0.78 & ---  & ---  & ASR \\
& \texttt{gpt-4o-mini-tts}     & ---  & ---  & 11.46 & 0.17 & TTS \\
\hdashline

\multirow{4}{*}{English} 
& \texttt{whisper-1}    & 0.03 & 0.01 & ---  & ---  & ASR \\
& \texttt{phi-4}               & 0.12 & 0.04 & ---  & ---  & ASR \\
& \texttt{gpt-4o-mini-tts}     & ---  & ---  & 11.44 & 0.06 & TTS \\
\hdashline

\multirow{4}{*}{Spanish} 
& \texttt{whisper-1}    & 0.02 & 0.01 & ---  & ---  & ASR \\
& \texttt{phi-4}               & 0.11 & 0.03 & ---  & ---  & ASR \\
& \texttt{gpt-4o-mini-tts}     & ---  & ---  & 10.84 & 0.07 & TTS \\

\bottomrule
\end{tabular}
}
}
\caption{Evaluation of ASR and TTS models across four languages. ASR performance is reported using Word Error Rate (WER) and Character Error Rate (CER). TTS is evaluated using Mel Cepstral Distortion (MCD) and CER, with the latter obtained via ASR using the \texttt{whisper-L-v3} model.}
\label{tab:asr_tts_appendix}
\end{table*}

\begin{table*}[!htbp]
\centering
\def\arraystretch{0.80}
{\scriptsize
\resizebox{0.8\textwidth}{!}{%
\begin{tabular}{l ccc ccc ccc c}
\toprule

\textbf{Language}& \textbf{R@1} & \textbf{R@5} & \textbf{R@10}
& \textbf{P@1} & \textbf{P@5} & \textbf{P@10}
& \textbf{F1@1} & \textbf{F1@5} & \textbf{F1@10}
&   \textbf{MRR}  \\ \midrule

\multicolumn{11}{c}{\cellcolor{Gray}{BM25 (T2T)}} \\ \midrule
Arabic & 13.26 & 35.57 & 45.12 & 15.65 & 8.55 & 5.45 & 14.04 & 13.55 & 9.62 & 25.25 \\
Chinese & 10.41 & 26.30 & 34.95 & 12.23 & 6.27 & 4.15 & 11.01 & 9.97 & 7.34 & 19.30 \\
English & 11.46 & 27.73 & 35.13 & 13.44 & 6.58 & 4.16 & 12.11 & 10.47 & 7.37 & 20.63 \\
Spanish & 13.58 & 33.78 & 43.37 & 15.88 & 8.02 & 5.17 & 14.34 & 12.75 & 9.15 & 24.93 \\
Average & 12.18 & 30.84 & 39.64 & 14.30 & 7.36 & 4.73 & 12.87 & 11.68 & 8.37 & 22.53 \\ \midrule

\multicolumn{11}{c}{\cellcolor{Gray}{\texttt{MiniLM-L12-v2} (T2T)}} \\ \midrule
Arabic & 7.13 & 24.76 & 36.12 & 8.82 & 6.09 & 4.46 & 7.68 & 9.60 & 7.85 & 16.87 \\
Chinese & 10.26 & 31.41 & 44.03 & 12.34 & 7.65 & 5.32 & 10.94 & 12.09 & 9.39 & 21.93 \\
English & 12.99 & 41.52 & 56.91 & 15.67 & 10.13 & 6.93 & 13.87 & 16.00 & 12.21 & 28.08 \\
Spanish & 10.70 & 34.20 & 48.56 & 13.07 & 8.40 & 5.94 & 11.48 & 13.23 & 10.47 & 23.64 \\
Average & 10.27 & 32.97 & 46.41 & 12.47 & 8.07 & 5.66 & 10.99 & 12.73 & 9.98 & 22.63 \\ \midrule

\multicolumn{11}{c}{\cellcolor{Gray}{\texttt{text-embedding-3L} (T2T)}} \\ \midrule
Arabic & 27.23 & 65.88 & 78.73 & 31.28 & 15.79 & 9.55 & 28.56 & 25.05 & 16.84 & 46.91 \\
Chinese & 29.89 & 70.63 & 83.11 & 34.47 & 16.93 & 10.06 & 31.39 & 26.85 & 17.75 & 50.93 \\
English & 32.58 & 75.72 & 88.03 & 37.87 & 18.22 & 10.72 & 34.31 & 28.87 & 18.89 & 54.80 \\
Spanish & 30.06 & 71.82 & 84.57 & 34.76 & 17.22 & 10.27 & 31.60 & 27.31 & 18.10 & 51.39 \\
Average & 29.94 & 71.01 & 83.61 & 34.59 & 17.04 & 10.15 & 31.46 & 27.02 & 17.89 & 51.01 \\ \midrule

\multicolumn{11}{c}{\cellcolor{Gray}{\texttt{gte-multilingual-base} (T2T)}} \\ \midrule
Arabic & 20.65 & 58.31 & 74.10 & 24.50 & 14.25 & 9.04 & 21.91 & 22.48 & 15.93 & 40.09 \\
Chinese & 23.37 & 63.49 & 79.55 & 27.48 & 15.34 & 9.65 & 24.71 & 24.28 & 17.02 & 43.80 \\
English & 29.91 & 68.89 & 82.47 & 34.51 & 16.59 & 9.99 & 31.41 & 26.27 & 17.63 & 50.38 \\
Spanish & 23.50 & 62.40 & 79.53 & 27.88 & 15.14 & 9.69 & 24.93 & 23.93 & 17.07 & 43.98 \\
Average & 24.36 & 63.27 & 78.91 & 28.59 & 15.33 & 9.59 & 25.74 & 24.24 & 16.91 & 44.56 \\
\midrule

\multicolumn{11}{c}{\cellcolor{Gray}{\texttt{NV-Embed-v2}} (T2T)} \\ \midrule
Arabic & 12.80 & 32.03 & 41.82 & 15.06 & 7.75 & 5.09 & 13.54 & 12.26 & 8.97 & 23.91 \\
Chinese & 22.75 & 59.87 & 74.59 & 27.24 & 14.56 & 9.10 & 24.22 & 23.01 & 16.03 & 42.47 \\
English & 24.65 & 70.28 & 87.35 & 29.45 & 17.16 & 10.68 & 26.22 & 27.09 & 18.82 & 47.48 \\
Spanish & 23.60 & 64.57 & 81.94 & 28.19 & 15.74 & 10.01 & 25.10 & 24.85 & 17.63 & 44.96 \\
Average & 20.95 & 56.68 & 71.42 & 24.99 & 13.80 & 8.72 & 22.27 & 21.80 & 15.36 & 39.70 \\
\midrule

\multicolumn{11}{c}{\cellcolor{Gray}{\texttt{CLAP}} (S2T)} \\ \midrule
Ara   & 0.03 & 0.20 & 0.49 & 0.04 & 0.06 & 0.07 & 0.03 & 0.09 & 0.12 & 0.15 \\
Chn   & 0.08 & 0.23 & 0.35 & 0.11 & 0.06 & 0.05 & 0.09 & 0.09 & 0.08 & 0.19 \\
Eng   & 0.10 & 0.52 & 0.95 & 0.11 & 0.12 & 0.11 & 0.10 & 0.19 & 0.19 & 0.34 \\
Esp   & 0.18 & 0.42 & 0.93 & 0.20 & 0.11 & 0.11 & 0.19 & 0.17 & 0.20 & 0.39 \\
Average   & 0.10 & 0.34 & 0.68 & 0.11 & 0.09 & 0.08 & 0.10 & 0.14 & 0.15 & 0.27 \\

\midrule

\multicolumn{11}{c}{\cellcolor{Gray}{\texttt{SpeechT5}} (S2T)} \\ \midrule
Ara   & 0.14 & 0.32 & 0.49 & 0.16 & 0.08 & 0.06 & 0.15 & 0.13 & 0.11 & 0.28 \\
Chn   & 0.06 & 0.35 & 0.61 & 0.07 & 0.08 & 0.07 & 0.07 & 0.13 & 0.13 & 0.23 \\
Eng   & 0.07 & 0.16 & 0.49 & 0.07 & 0.04 & 0.06 & 0.07 & 0.06 & 0.10 & 0.16 \\
Esp   & 0.05 & 0.29 & 0.68 & 0.05 & 0.08 & 0.08 & 0.05 & 0.12 & 0.15 & 0.22 \\
Average   & 0.08 & 0.28 & 0.57 & 0.09 & 0.07 & 0.07 & 0.08 & 0.11 & 0.12 & 0.22 \\

\bottomrule
\end{tabular}%
}}
\caption{Knowledge retrieval performance in two settings: text-to-text retrieval using multilingual text encoders (T2T) and speech-to-text retrieval using multilingual and multimodal encoders (S2T). Retrieval is performed over a fully parallel set of knowledge snippets, enabling direct cross-lingual comparison. The T2T setting uses human-annotated transcripts, while S2T is based on user-recorded audio.}
\label{tab:retrieval_appendix}
\end{table*}

\begin{table}[!htbp]
\centering
\def\arraystretch{0.80}
{\scriptsize
\resizebox{0.8\columnwidth}{!}{%
\begin{tabular}{l ccc}
\toprule
\textbf{Language} & \textbf{Turn Acc.} & \textbf{Exact Match} & \textbf{OOK Recall} \\ \midrule

\multicolumn{4}{c}{\cellcolor{Gray}{\texttt{XLM-R\textsubscript{large}}}} \\ \midrule
Arabic   & 95.39 & --- & --- \\
Chinese  & 95.23 & --- & --- \\
English  & 96.30 & --- & --- \\
Spanish  & 95.93 & --- & --- \\
Average  & 95.71 & --- & --- \\ \midrule

\multicolumn{4}{c}{\cellcolor{Gray}{\texttt{Llama-3.1-8B-Inst} (R@5)}} \\ \midrule
Arabic   & 86.75 & 15.26 & 5.06 \\
Chinese  & 92.99 & 20.73 & 0.00 \\
English  & 92.19 & 18.14 & 6.33 \\
Spanish  & 93.30 & 21.00 & 13.92 \\
Average  & 91.31 & 18.78 & 6.83 \\ \midrule

\multicolumn{4}{c}{\cellcolor{Gray}{\texttt{gpt-4.1-nano}} (R@5)} \\ \midrule
Arabic   & --- & 19.96 (20.21$^{*}$) & 3.80 (0.00$^{*}$) \\
Chinese  & --- & 22.98 (19.86$^{*}$) & 2.53 (0.00$^{*}$)  \\
English  & --- & 26.62  (23.02$^{*}$) & 5.06 (14.29$^{*}$) \\
Spanish  & --- & 24.87 (21.09$^{*}$) & 5.06 (0.00$^{*}$) \\
Average  & --- & 23.61 (21.05$^{*}$) & 4.11 (3.57$^{*}$) \\ 
\midrule

\multicolumn{4}{c}{\cellcolor{Gray}{\texttt{gpt-4.1-nano}} (R@10)} \\ \midrule
Arabic   & --- & 12.58 & 3.80 \\
Chinese  & --- & 17.15 & 0.00 \\
English  & --- & 23.33 & 1.27 \\
Spanish  & --- & 19.55 & 3.80 \\
Average  & --- & 18.15 & 2.22 \\ 
\midrule

\multicolumn{4}{c}{\cellcolor{Gray}{\texttt{gpt-4o-nano}} (R@20)} \\ \midrule
Arabic   & --- & 10.85 & 1.27 \\
Chinese  & --- & 12.28 & 0.00 \\
English  & --- & 18.72 & 1.27 \\
Spanish  & --- & 11.03 & 0.00 \\
Average  & --- & 13.72 & 0.63 \\ 
\midrule

\multicolumn{4}{c}{\cellcolor{Gray}{Threshold}} \\ \midrule
Arabic   & --- & 6.26 & 48.47 \\
Chinese  & --- & 6.61 & 46.38 \\
English  & --- & 6.88 & 63.10 \\
Spanish  & --- & 6.46 & 43.70 \\
Average  & --- & 6.55 & 50.41 \\ 
\midrule

\multicolumn{4}{c}{\cellcolor{Gray}{\texttt{gpt-4.1}} (R@5)} \\ \midrule
Arabic   & --- & 34.27$^{*}$ & 0.00$^{*}$ \\
Chinese  & --- & 39.19$^{*}$ & 14.29$^{*}$ \\
English  & --- & 44.29$^{*}$ & 42.86$^{*}$ \\
Spanish  & --- & 39.54$^{*}$ & 14.29$^{*}$ \\
Average  & --- & 39.32$^{*}$ & 17.36$^{*}$ \\ 
\midrule

\multicolumn{4}{c}{\cellcolor{Gray}{\texttt{gpt-4.1-mini}} (R@5)} \\ \midrule
Arabic   & --- & 14.94$^{*}$ & 0.00$^{*}$ \\
Chinese  & --- & 23.02$^{*}$ & 14.29$^{*}$ \\
English  & --- & 26.19$^{*}$ & 0.00$^{*}$ \\
Spanish  & --- & 20.56$^{*}$ & 0.00$^{*}$ \\
Average  & --- & 21.18$^{*}$ & 3.57$^{*}$ \\ 
\midrule

\multicolumn{4}{c}{\cellcolor{Gray}{\texttt{gpt-4o}} (R@5)} \\ \midrule
Arabic   & --- & 30.05$^{*}$ & 14.29$^{*}$ \\
Chinese  & --- & 35.32$^{*}$ & 42.86$^{*}$ \\
English  & --- & 36.91$^{*}$ & 28.57$^{*}$ \\
Spanish  & --- & 33.04$^{*}$ & 42.86$^{*}$ \\
Average  & --- & 33.83$^{*}$ & 32.14$^{*}$ \\ 
\midrule

\multicolumn{4}{c}{\cellcolor{Gray}{\texttt{gpt-4o-mini}} (R@5)} \\ \midrule
Arabic   & --- & 7.91$^{*}$ & 0.00$^{*}$ \\
Chinese  & --- & 11.78$^{*}$ & 0.00$^{*}$ \\
English  & --- & 10.37$^{*}$ & 0.00$^{*}$ \\
Spanish  & --- & 10.02$^{*}$ & 0.00$^{*}$ \\
Average  & --- & 10.02$^{*}$ & 0.00$^{*}$ \\

\bottomrule
\end{tabular}%
}}

\caption{Model performance on retrieval turn classification and knowledge filtering across four languages. Turn classification is evaluated using accuracy (Turn Acc.). Knowledge filtering is measured by Exact Match (EM) and Out-of-Knowledge (OOK) Recall. R@5, R@10, and R@20 indicate the number of top-ranked retrieved snippets considered during filtering. The upper bounds for EM under R@5, R@10, and R@20 are 72.53\%, 86.41\%, and 92.26\%, respectively (averaged across all languages). ($^*$) Results marked with an asterisk are based on the same randomly sampled 10\% subset of the test set, due to the high cost of evaluating the full dataset with these models. This subset contains only 7 OOK turns, leading to high variance in OOK Recall, and should be interpreted with caution.}
\label{tab:knowledge_selection_appendix}
\end{table}

\begin{table}[!htbp]
\centering
\def\arraystretch{0.80}
{\scriptsize
\resizebox{0.9\columnwidth}{!}{%
\begin{tabular}{l ccc}
\toprule
\textbf{Language} & \textbf{BLEU} & \textbf{METEOR} & \textbf{ROUGE-L} \\ \midrule

\multicolumn{4}{c}{\cellcolor{Gray}{\texttt{Llama-3.1-8B-Inst}}} \\ \midrule
Arabic   & 3.35 & 15.11 & 18.24 \\
Chinese  & 4.86 & 28.20 & 0.01 \\
English  & 5.67 & 31.45 & 26.98 \\
Spanish  & 5.76 & 26.28 & 26.22 \\
Average  & 4.91 & 25.26 & 17.86 \\ \midrule

\multicolumn{4}{c}{\cellcolor{Gray}{\texttt{gpt-4o-nano}}} \\ \midrule
Arabic   & 3.85 & 22.93 & 20.85 \\
Chinese  & 5.96 & 33.76 & 0.00 \\
English  & 10.99 & 55.65 & 36.67 \\
Spanish  & 6.57 & 32.77 & 29.59 \\
Average  & 6.84 & 36.28 & 21.78 \\ 
 \bottomrule
\end{tabular}%
}}

\caption{Model performance on response generation. At time step $t$, the model receives as input the dialogue history $\mathcal{H}_t$, the ground-truth set of knowledge snippets $\mathcal{K}_t$, and a retrieval indicator $\mathbf{r}_t$, and generates a system response $\hat{\mathbf{s}}_t$.}
\label{tab:response_generation_appendix}
\end{table}

\end{document}